\documentclass[letterpaper]{article} 
\usepackage[preprint]{aaai2027}
\copyrighttext{\textit{Preprint}. Copyright \copyright\ 2026 by the authors.}
\usepackage[hyphens]{url}  
\usepackage{graphicx} 
\usepackage{natbib}  
\usepackage{caption} 
\usepackage{algorithm}
\usepackage{algorithmic}

\usepackage{newfloat}
\usepackage{listings}
\DeclareCaptionStyle{ruled}{labelfont=normalfont,labelsep=colon,strut=off} 
\floatstyle{ruled}
\newfloat{listing}{tb}{lst}{}
\floatname{listing}{Listing}

\usepackage{booktabs}
\usepackage{mathtools}
\usepackage{amssymb}
\definecolor{linkorange}{RGB}{211,84,0}
\usepackage[
  colorlinks=true,
  linkcolor=linkorange,
  citecolor=linkorange,
  urlcolor=linkorange
]{hyperref}
\newsavebox{\bestnormalbox}
\newsavebox{\bestboldbox}
\newcommand{\bestcell}[1]{%
  \sbox{\bestnormalbox}{#1}%
  \sbox{\bestboldbox}{{\boldmath#1}}%
  \resizebox{\the\wd\bestnormalbox}{\the\ht\bestboldbox}%
    {{\boldmath#1}}%
}
\renewcommand{\underline}[1]{\bestcell{#1}}
\newtheorem{proposition}{Proposition}

\title{Diagnose, Recover, Certify:\\Task Readiness under Hidden Dynamics Changes}
\author{Nguyen Viet Tuan Kiet$^1{}^\dagger$, Huynh Thi Thanh Binh$^1{}^\dagger$}
\affiliations{$^1$Hanoi University of Science and Technology, Hanoi, Vietnam\\${}^\dagger$Corresponding authors: \texttt{tuankiet.nv2501@gmail.com, binhht@soict.hust.edu.vn} }

\begin{document}

\maketitle

\begin{abstract}
A deployed control policy can conceal consequential dynamics changes: an actuator may lose effectiveness without affecting the current task when the policy rarely excites it, despite being critical for a future task that has not yet been specified. We introduce task readiness under dormant dynamics drift, a decision problem that unifies active change diagnosis and post-change control recovery under a limited, task-agnostic interaction budget. An agent must identify whether and where local dynamics have changed, use a small number of informative interactions to characterize the change before downstream task identity is revealed, and subsequently provide each candidate task with either a recovered policy and a calibrated lower bound on its achievable return or an abstention decision to a safe fallback. We propose Evidence-Gated Matched-Pulse Transport, an intervention-based Bayesian procedure that couples fault localization with estimation of actuator effectiveness through a shared matched-response representation, thereby preserving diagnostic reliability while converting localized evidence into recovery-relevant uncertainty. This uncertainty is propagated to task-conditioned policy selection and readiness certification, enabling deployment decisions that explicitly trade off expected performance, confidence, and fallback use. We evaluate the resulting framework on a diverse suite of dormant-actuator benchmarks spanning multiple simulators, under a protocol that separates diagnosis from capability recovery, scores deployment by readiness coverage, selective risk, and interaction cost as well as return, and identifies the fault regimes in which transported evidence is decisive.
\end{abstract}


\section{Introduction}

Reinforcement-learning agents are increasingly expected to operate after their
environment has changed, not only during the stationary training regime in
which their policies were learned. A robot can lose effectiveness in one joint,
a vehicle can suffer degradation in a control surface, and an industrial system
can develop a fault in a redundant actuator. Such changes need not immediately
degrade the task currently being executed. If the deployed policy never
substantially excites the affected mechanism, its observations can remain
indistinguishable from those of the nominal system, even though a later task
will depend critically on the missing capability for successful execution.

This situation is not well captured by the usual view of non-stationary RL.
Methods for continual or structured non-stationarity adapt from a changing
stream of trajectories \citep{feng2022fans,xie2021continual}, while
changepoint approaches seek to identify a shift from its sequential evidence
and control false alarms or detection delay \citep{alami2023rbocpd,liu2024bada}.
Both are valuable when a changed reward or transition kernel leaves a footprint
in the data observed by the current policy. Here, however, passive evidence can
be exactly uninformative: the nominal and changed systems induce the same
trajectory law until the agent chooses an intervention that activates the
dormant mechanism. The resulting challenge is therefore one of observability,
not merely limited data or slow adaptation.

Active identification offers the natural next step, but does not by itself
resolve the decision problem. Probing and exploration methods can infer latent
dynamics before transfer \citep{yang2020sept}, explore uncertain systems for
downstream planning \citep{sukhija2023opax}, and allocate experiments according
to their control value \citep{wagenmaker2023optimal,memmel2024asid}. Meta-RL
and off-dynamics transfer likewise use limited interaction to infer a context
or adapt across mismatched dynamics \citep{rakelly2019pearl,lyu2024odrl,
liu2024offdynamics}. These approaches generally prepare for a known task,
known task distribution, or target environment. They do not isolate the case in
which a local change must first be diagnosed, recovery data must be collected
before the next task is specified, and the agent must subsequently decide for
each candidate task whether deployment is justified.

Safety and uncertainty-aware decision-making address a complementary part of
this problem. Conservative policy-improvement methods limit unsupported updates
\citep{laroche2019spibb}, while safe model-based control and sampling-based
safe RL protect learning and deployment with stability or constraint guarantees
\citep{berkenkamp2017safe,suttle2024sampling}. Selective prediction formalizes
the option to abstain when uncertainty is high \citep{geifman2019selectivenet}.
Yet these lines do not provide an end-to-end mechanism that connects
hidden-fault diagnosis to capability recovery and then to a task-wise
deploy-or-abstain certificate. In particular, they do not answer which
interventions should be taken when the future task is still unknown, or how
diagnostic evidence should be preserved without allowing a misspecified
severity model to corrupt fault localization.

We formulate this setting as \emph{task readiness under dormant dynamics
drift}. Before a future task is revealed, an agent receives a limited,
task-agnostic interaction budget. It must determine whether and where the
dynamics changed, estimate the resulting loss of effectiveness sufficiently
well to recover a family of task policies, and output for each task either a
lower-bound readiness assessment or an abstention to a fallback. This
formulation separates two decisions that are often conflated: locating the
changed mechanism and estimating its magnitude for control.

To solve this problem, we introduce \emph{Evidence-Gated Matched-Pulse
Transport} (EG-MPT). EG-MPT uses matched-response scores from targeted
diagnostic trajectories to maintain a categorical posterior over fault
locations. A continuous response coordinate is transported to a conditional
severity posterior only when categorical evidence favors the corresponding
fault. This design retains amplitude information for recovery while making the
location belief, acquisition sequence, and alarm path invariant to
severity-model error. The recovered posterior is then propagated through
task-conditioned policy correction and a posterior-predictive lower confidence
bound, yielding an operational readiness decision under a fixed interaction
budget.

\paragraph{Contributions.}
Our contributions are threefold:
\begin{enumerate}
    \item We formulate \emph{task readiness under dormant dynamics drift}:
    a component-level change is hidden by the deployed policy, recovery data
    precede future-task reveal, and deployment requires a task-wise
    recover-or-abstain decision.  The formulation separates fault localization
    from effectiveness estimation under a fixed budget.
    \item We propose \emph{Evidence-Gated Matched-Pulse Transport} (EG-MPT),
    which uses one targeted response for categorical localization and gated
    effectiveness transport.  We prove pathwise invariance of the
    localization posterior, diagnostic sequence, and stopping decision to arbitrary
    severity-model error; under the calibrated joint model, the transported
    coordinate weakly improves fixed-budget Bayes control risk and readiness
    value, with a quantitative control-value bound.
    \item We evaluate the protocol on a diverse suite of dormant-actuator
    benchmarks spanning multiple simulators and dynamics families, comparing
    against adapted published methods and matched controls under
    fault-severity, recovery-budget, and deadline stress.  Beyond detection, we
    measure effectiveness error, deployed return, certificate violations, and
    interaction cost, identifying when diagnostic evidence yields future-task
    capability.
\end{enumerate}

\section{Related Work}
\label{sec:related_work}

\subsection{Non-Stationary Reinforcement Learning}

Non-stationary RL studies adaptation to changing transitions or rewards through
localized change representations, regret objectives, and changepoint-aware
updates \citep{feng2022fans,chen2022goaloriented,liu2024bada}. Off-dynamics RL
instead transfers policies across mismatched source and target dynamics using
benchmarks, robust objectives, or limited target-domain modeling
\citep{lyu2024odrl,liu2024offdynamics,guo2026mobody}, while active
quickest-change methods select costly sensing channels to balance detection
delay and false alarms \citep{gopalan2021bandit}. These approaches typically
act once a shift affects observed behavior or treat detection and transfer as
separate objectives. Our setting isolates a sharper observability problem: a
component change can induce exactly the nominal trajectory law under the
deployed policy. Targeted interventions must therefore reveal the change while
also producing evidence for task-agnostic recovery and readiness assessment
before the future task is known. Appendix~\ref{app:related_work} provides a
more detailed comparison across these neighboring research directions.

\subsection{Active Identification and Safe Transfer}

Active identification uses probing and experiment design to acquire dynamics
information that matters for subsequent control. SEPT infers latent dynamics
before policy execution, OPAX explores for downstream planning, and
task-oriented design prioritizes parameters by control value
\citep{yang2020sept,sukhija2023opax,wagenmaker2023optimal}; ASID similarly
designs informative real-world exploration from an initial simulator
\citep{memmel2024asid}. Safe-transfer methods complement this objective by
limiting unsupported policy updates or enforcing constraints during learning
and deployment \citep{laroche2019spibb,suttle2024sampling}. Our problem joins
these concerns in a different sequence: first locate a change hidden by the
current policy, then spend a small task-agnostic budget before task reveal, and
finally issue a task-wise deploy-or-abstain decision. Appendix~\ref{app:related_work}
discusses these connections and their differing assumptions in greater detail.

\section{Problem Formulation}
\label{sec:problem_formulation}

\subsection{Task-Readiness Setting}

At time $t$, the agent observes a state $s_t\in\mathcal{S}$, selects an action
$a_t\in\mathcal{A}$, and receives the next state according to
$s_{t+1}\sim P_{H,\mathbf{g}}(\cdot\mid s_t,a_t)$, where
$H\in\mathcal{H}:=\{0,1,\ldots,m\}$ identifies the location of a possible
change and $\mathbf{g}=(g_1,\ldots,g_m)$ describes the effectiveness of the
system's $m$ mechanisms. We assume
$\mathbf{g}\in[g_{\min},1]^m$ for a known $g_{\min}>0$. The nominal system has
$H=0$ and $\mathbf{g}=\mathbf{1}$, where $\mathbf 1$ is the all-ones
effectiveness vector; $H=i$ indicates that mechanism $i$ changes at an unknown
time $\tau$, with $g_k=1$ for every $k\neq i$. Thus at most one mechanism
changes.

Let $\mathcal{K}$ be the set of possible future tasks. A task
$\kappa\sim\nu$ is drawn from distribution $\nu$ but revealed only at time
$R>\tau$. Each task has reward function $r_\kappa$, and the discounted return
of policy $\pi$ with discount factor $\gamma\in[0,1)$ is
\begin{equation}
    J_\kappa(\pi;H,\mathbf{g})
    :=
    \mathbb{E}_{\pi;\,P_{H,\mathbf{g}}}
    \left[
        \sum_{t=0}^{\infty}
        \gamma^t r_\kappa(s_t,a_t)
    \right].
    \label{eq:task_return}
\end{equation}
Before $R$, the agent may collect only task-agnostic data and must prepare a
policy $\pi_\kappa$ and readiness assessment for every candidate task.  After
reveal, it deploys $\pi_\kappa$ or abstains to a designated fallback
$\pi_\kappa^{\mathrm{fallback}}$.

\subsection{Dormant Dynamics Drift}

The difficulty is practical whenever the current policy $\pi_0$ activates only
a subset of the mechanisms.  A robot may leave some joints unused, a vehicle
may keep a control surface inactive, or an industrial system may hold a
redundant actuator in reserve.  Such a mechanism can degrade without changing
current observations yet become essential after task reveal; exhaustive
testing is costly, while waiting until reveal leaves little time for recovery.

Formally, let
$\mathbb{P}^{\pi_0}_{H,\mathbf{g},T}$ denote the distribution of passive
trajectories up to time $T$.  We call a change in mechanism $i$ \emph{dormant
under $\pi_0$} when
\begin{equation}
    \mathbb{P}^{\pi_0}_{0,\mathbf{1},T}
    =
    \mathbb{P}^{\pi_0}_{i,\mathbf{g},T},
    \quad
    \forall T\in\{\tau,\ldots,R-1\},
    \label{eq:dormant_equivalence}
\end{equation}
where $\mathbf{g}$ differs from $\mathbf{1}$ only at coordinate $i$.
Equality in Eq.~\eqref{eq:dormant_equivalence} implies zero passive
information about the change; targeted intervention is necessary.  Sufficient
occupancy conditions and actuator examples are given in
Appendix~\ref{app:formal_problem_details}.

\subsection{Two-Stage Readiness Protocol}

The agent receives task-agnostic budgets $B_1$ and $B_2$.  Stage~1 selects
complete diagnostic trajectories from a finite library $\mathcal Q$ to detect
and localize a change.  Stage~2 uses a small number of recovery trajectories to
characterize its magnitude and refine joint uncertainty over $(H,\mathbf g)$,
yielding terminal belief $b^{(2)}$.  This separation matters because a correct
location alone does not determine how much control authority remains, while
recovery data must be collected before the future task is known.  The
intervention and posterior interfaces are detailed in
Appendix~\ref{app:formal_problem_details}; Section~\ref{sec:methodology}
specifies how EG-MPT realizes these interfaces in its end-to-end procedure.

For every $\kappa\in\mathcal K$, the output is a recovered policy
$\pi_\kappa$ and continuous lower return estimate $L_\kappa$.  At error level
$\alpha\in(0,1)$, the intended certificate semantics are
\begin{equation}
    \Pr\!\left(
        J_\kappa(\pi_\kappa;H,\mathbf{g})
        \geq L_\kappa
        \,\middle|\,
        b^{(2)}
    \right)
    \geq 1-\alpha,
    \qquad
    \forall \kappa\in\mathcal{K}.
    \label{eq:readiness_certificate}
\end{equation}
conditional on the modeled change family and calibration assumptions.  Given
$J_{\min}$, the deployed policy $\pi_\kappa^{\mathrm{dep}}$ equals
$\pi_\kappa$ when $L_\kappa\geq J_{\min}$ and the fallback otherwise.  The
principal objective is expected deployed regret plus interaction cost:
\begin{equation}
    \mathcal{R}_{\mathrm{dep}}
    :=
    \mathbb{E}_{\kappa\sim\nu}
    \left[
        J_\kappa^\star(H,\mathbf{g})
        -
        J_\kappa(
            \pi_\kappa^{\mathrm{dep}};
            H,\mathbf{g}
        )
    \right],
    \label{eq:end_to_end_regret}
\end{equation}
where $J_\kappa^\star(H,\mathbf g):=\sup_\pi
J_\kappa(\pi;H,\mathbf g)$.  Evaluation metrics and their estimators are
specified in Appendix~\ref{app:experimental_details}.

\section{Methodology}
\label{sec:methodology}

\begin{figure*}[t]
    \centering
    \includegraphics[width=0.9\linewidth]{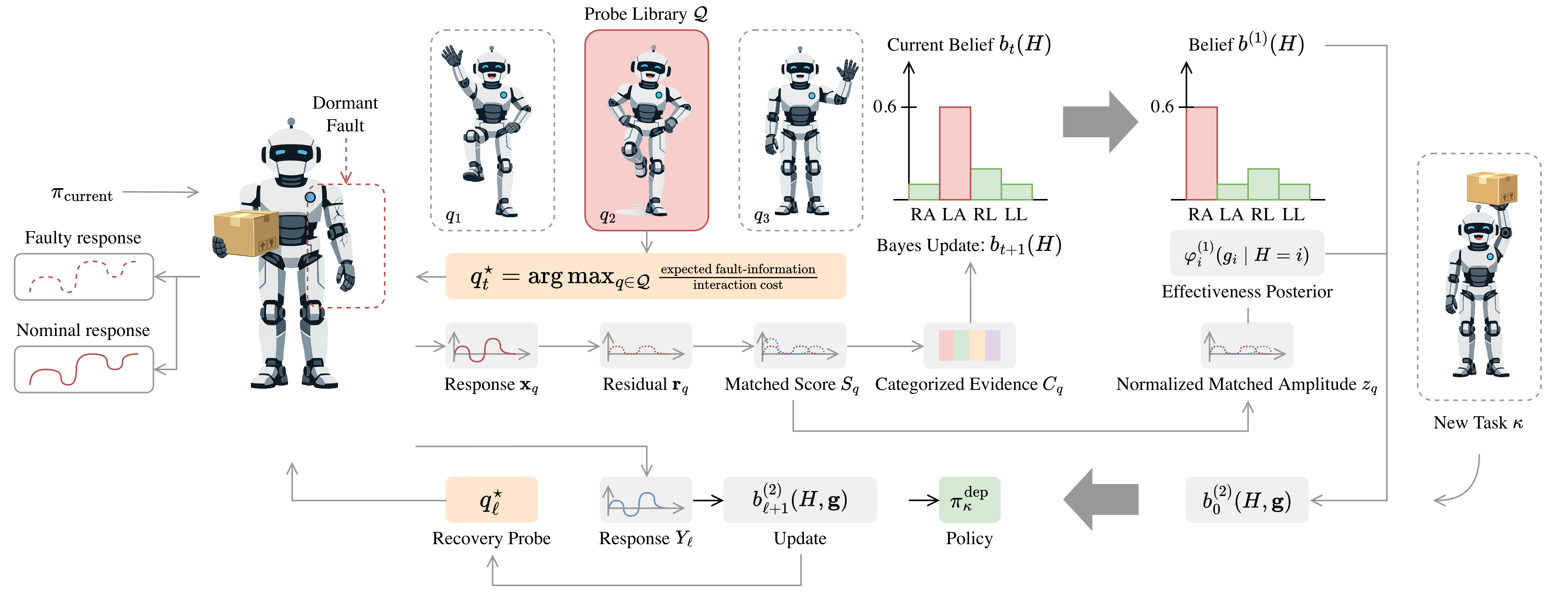}
    \caption{Overview of Evidence-Gated Matched-Pulse Transport (EG-MPT).
    Stage~1 selects task-agnostic diagnostic probes, converts their responses
    into categorized matched-score evidence for Bayesian fault localization,
    and gates the corresponding normalized matched amplitude into an
    actuator-effectiveness posterior.  Stage~2 initializes recovery from this
    transported joint belief, selects recovery probes, and produces a
    task-conditioned deployment policy after the new task is revealed.}
    \label{fig:eg_mpt_overview}
\end{figure*}

\emph{Evidence-Gated Matched-Pulse Transport} (EG-MPT), illustrated in
Figure~\ref{fig:eg_mpt_overview}, separates robust fault localization from the
continuous severity information needed for recovery.  At
Stage~1 round $t$, let $\mathcal D_t$ contain the passive observations and
quantized diagnostic outcomes collected so far, and define
$b_t(h):=\Pr(H=h\mid\mathcal D_t)$.  A separate conditional posterior
$\varphi_{t,i}(g):=p(g_i=g\mid H=i)$ uses only continuous evidence admitted by
the gate below; it never enters Stage~1 acquisition or stopping.  If
$b^{(1)}$ and $\varphi_i^{(1)}$ are their terminal values, Stage~2 begins from
\begin{equation}
    b^{(2)}_0(H,\mathbf g)
    =b^{(1)}(H)p_0(\mathbf g\mid H),
    \label{eq:recovery_initialization}
\end{equation}
with $p_0$ instantiated from the transported conditionals below.  The same
diagnostic response thus supports both stages without coupling their error
models.

\subsection{Matched-Response Localization}
\label{sec:method_location}

Let $j(q)\in\{1,\ldots,m\}$ be the mechanism excited by diagnostic trajectory
$q$, and let $\mathbf x_q$ be its stacked observed response.  From nominal and
calibrated-fault predictions at $g_{\mathrm{cal}}\in[g_{\min},1)$, define
\begin{equation}
    \mathbf{r}_q
    :=
    \mathbf{x}_q-\widehat{\mathbf{x}}_q^{\,0},
    \qquad
    \mathbf{h}_q
    :=
    \widehat{\mathbf{x}}_q^{\,j(q),g_{\mathrm{cal}}}
    -\widehat{\mathbf{x}}_q^{\,0}.
    \label{eq:residual_signature}
\end{equation}
for trajectories with $\lVert\mathbf h_q\rVert_2>0$.  The matched-response
score
\begin{equation}
    S_q
    :=
    \frac{\langle\mathbf{r}_q,\mathbf{h}_q\rangle}
         {\lVert\mathbf{h}_q\rVert_2}.
    \label{eq:matched_score}
\end{equation}
retains signed evidence in the predicted fault direction with a noise scale
that does not grow with signature magnitude \citep{turin1960matched}.
Held-out calibration quantizes $S_q$ into $C_q\in\{1,\ldots,K\}$ and estimates
$p_q(c\mid h):=\Pr(C_q=c\mid H=h,q)$.  The location update is
\begin{equation}
    b_{t+1}(h)
    =
    \frac{p_q(C_q\mid h)b_t(h)}
         {\sum_{h'\in\mathcal{H}}p_q(C_q\mid h')b_t(h')}.
    \label{eq:stage1_location_update}
\end{equation}
after a change-hazard prediction.  Writing
$p(c\mid b_t,q)=\sum_h p_q(c\mid h)b_t(h)$ and $b_{t+1}^{q,c}$ for the
hypothetical posterior, EG-MPT selects the greatest task-weighted Bayes-risk
reduction per interaction charge $\omega_q>0$:
\begin{equation}
    q_t^\star
    \in
    \arg\max_{q\in\mathcal{Q}}
    \frac{
        \rho(b_t)
        -
        \mathbb{E}_{C_q\sim p(\cdot\mid b_t,q)}
        [\rho(b_{t+1}^{\,q,C_q})]
    }{
        \omega_q
    },
    \label{eq:stage1_acquisition}
\end{equation}
where $\rho$ is a fixed nonnegative Bayes-risk functional calibrated from the
future-task distribution $\nu$.

Because severity never enters
Eqs.~\eqref{eq:stage1_location_update}--\eqref{eq:stage1_acquisition}, two
executions differing only in their severity update obey the following result;
primed symbols denote the second execution.

\begin{proposition}[Severity-orthogonal localization]
\label{prop:severity_orthogonality}
Fix the categorical channel, initialization, stopping rule, tie-breaking rule,
and exogenous random stream.  For any two, possibly misspecified, severity
updates and any Stage~1 horizon $T$,
\begin{equation}
    \sup_{t\leq T}\lVert b_t-b'_t\rVert_{\mathrm{TV}}=0,
    \qquad q_t^\star={q_t^\star}'\quad\text{a.s.}
    \label{eq:severity_orthogonality}
\end{equation}
Their stopping decisions also coincide.  Consequently, localization loss,
detection time, and Stage~1 interaction charge are invariant to arbitrary
severity-model error.
\end{proposition}

\subsection{Evidence-Gated Transport and Recovery}
\label{sec:method_transport}

Quantization is useful for robust localization but discards within-bin
amplitude.  EG-MPT therefore retains one normalized matched coordinate:
\begin{equation}
    z_q
    :=
    \frac{
        S_q/\lVert\mathbf h_q\rVert_2-m_{0,j(q)}
    }{
        m_{1,j(q)}-m_{0,j(q)}
    },
    \label{eq:matched_coordinate}
\end{equation}
where $m_{0,j}$ and $m_{1,j}$ are robust centers of the least-squares
coefficient under nominal dynamics and a calibrated fault, respectively.
Here the second division is intentional:
$S_q/\lVert\mathbf h_q\rVert_2$ is the least-squares signature coefficient.
The outer normalization places nominal and calibrated-fault responses near
$0$ and $1$, so locally $z_q\approx(1-g)/(1-g_{\mathrm{cal}})$.  Its robust
dispersion $\sigma_j>0$ is estimated on the same calibration split.

The categorical channel decides whether the continuous coordinate is allowed
to affect severity.  For a trajectory targeting $j=j(q)$, the gate opens
exactly when
\begin{equation}
    p_q(C_q\mid j)>p_q(C_q\mid0).
    \label{eq:evidence_gate}
\end{equation}
This positive-likelihood-ratio region maximizes, over deterministic gates, the
difference between its opening probabilities under $H=j$ and $H=0$
\citep{neyman1933efficient}.  A closed gate leaves $\varphi_{t,j}$ unchanged;
an open gate updates it on a gain support
$\mathcal{G}\subset[g_{\min},1]$, where $\mathcal{G}$ is a finite grid that
contains both $1$ and $g_{\mathrm{cal}}$.  For every $g\in\mathcal G$,
\begin{equation}
    \varphi_{t+1,j}(g)
    \propto
    \varphi_{t,j}(g)
    \exp\!\left[
        -\frac{
            \left(z_q-\frac{1-g}{1-g_{\mathrm{cal}}}\right)^2
        }{2\sigma_j^2}
    \right],
    \label{eq:gated_gain_update}
\end{equation}
This conditional update cannot feed back into $b_t$; the gate prevents
nominal-supporting fluctuations from fabricating a confident severity.

At the Stage~1--Stage~2 boundary, let $\varphi_i^{(1)}$ denote the final
conditional posterior for mechanism $i$.  EG-MPT instantiates the prior in
Eq.~\eqref{eq:recovery_initialization} as
\begin{equation}
    \begin{aligned}
        p_0(\mathbf{g}\mid H=0)
        &=
        \prod_{k=1}^{m}\delta_1(g_k),\\
        p_0(\mathbf{g}\mid H=i)
        &=
        \varphi_i^{(1)}(g_i)
        \prod_{k\neq i}\delta_1(g_k),
    \end{aligned}
    \label{eq:transported_initialization}
\end{equation}
where $\delta_1$ is a point mass at nominal effectiveness.  Thus the
Stage~2 state continues the uncertainty accumulated during localization for
downstream recovery.

For a Stage~2 response $Y_\ell$ to trajectory $q_\ell$, the joint state is
updated by
\begin{equation}
    b_{\ell+1}^{(2)}(H,\mathbf g)
    \propto
    p(Y_\ell\mid H,\mathbf g,q_\ell)b_\ell^{(2)}(H,\mathbf g).
    \label{eq:stage2_gain_likelihood}
\end{equation}
We implement this likelihood with counterfactual trajectory discrepancies;
Appendix~\ref{app:formal_problem_details} gives the ensemble form and
normalization details.

Recovery acquisition minimizes future control loss rather than gain variance
in isolation.  Let $G_i$ denote the random effectiveness of mechanism $i$
under $b_\ell^{(2)}$, with $G_i=1$ whenever $H\neq i$.  Let $w_i\geq0$ be a
fixed importance weight calibrated from the future-task distribution $\nu$.
EG-MPT selects
\begin{equation}
    q_\ell^\star
    \in
    \arg\max_{q\in\mathcal{Q}}
    \frac{
        w_{j(q)}\left(1-
        \frac{
            \mathbb E_{b_\ell^{(2)}}[G_{j(q)}]^2
        }{
            \mathbb E_{b_\ell^{(2)}}[G_{j(q)}^2]
        }\right)
    }{
        \omega_q
    }.
    \label{eq:stage2_acquisition}
\end{equation}
The numerator is the residual risk after the best scalar gain correction, so
recovery focuses on uncertainty that matters for future control.

To isolate the value of transport, let $\mathcal F_0$ contain the categorical
Stage~1 history and any common recovery observations, and let
$\mathcal F_1$ additionally contain $z_q$ whenever the gate opens.  Thus
$\mathcal F_0\subseteq\mathcal F_1$.  For either information set
$\mathcal F$, define
\begin{equation}
    c_i(\mathcal F)
    :=
    \frac{
        \mathbb{E}[G_i\mid\mathcal{F}]
    }{
        \mathbb{E}[G_i^2\mid\mathcal{F}]
    },
    \quad
    R_i(\mathcal F)
    :=
    \mathbb{E}[(G_i c_i(\mathcal F)-1)^2].
    \label{eq:information_specific_risk}
\end{equation}
\begin{proposition}[Quantified control value]
\label{prop:control_value}
Under the calibrated gain model, the transported coordinate cannot increase
control risk and, more precisely,
\begin{align}
    g_{\min}^2
    \mathbb{E}\!\left[
        \operatorname{Var}
        (c_i(\mathcal F_1)\mid\mathcal F_0)
    \right]
    &\leq
    R_i(\mathcal F_0)-R_i(\mathcal F_1)
    \nonumber\\
    &\leq
    \mathbb{E}\!\left[
        (c_i(\mathcal F_0)-c_i(\mathcal F_1))^2
    \right].
    \label{eq:control_value_bounds}
\end{align}
The improvement is strict if and only if the gated coordinate changes the
posterior-optimal correction with positive probability under the induced data
distribution.
\end{proposition}

Thus the transported coordinate refines the categorical experiment in the
decision-theoretic sense \citep{blackwell1953equivalent}; the proof and strictness
conditions are in Appendix~\ref{app:theory}.

\subsection{Recovery and Certification}
\label{sec:method_certificate}

After the recovery budget is exhausted, EG-MPT converts $b^{(2)}$ into the
policy family.  For each mechanism, the Bayes correction under squared
actuation mismatch is
\begin{equation}
    c_i^\star
    :=
    \arg\min_c
    \mathbb{E}_{b^{(2)}}[(G_i c-1)^2]
    =
    \frac{
        \mathbb{E}_{b^{(2)}}[G_i]
    }{
        \mathbb{E}_{b^{(2)}}[G_i^2]
    }.
    \label{eq:bayes_gain_correction}
\end{equation}
The recovered policy applies these corrections to $\pi_\kappa^0$ and clips
actions; marginalizing both $H$ and $G_i$ avoids a hard location decision.
The joint posterior and world-model ensemble induce a predictive return
$\widehat J_\kappa$ with mean $\widehat\mu_\kappa$ and variance
$\widehat\sigma_\kappa^2$.  A one-sided Cantelli bound gives
\begin{equation}
    L_\kappa
    :=
    \widehat{\mu}_\kappa
    -
    \sqrt{
        \frac{1-\alpha}{\alpha}
        \widehat{\sigma}_\kappa^2
    }.
    \label{eq:cantelli_certificate}
\end{equation}
Under a calibrated posterior predictive model,
$\Pr(\widehat{J}_\kappa\geq L_\kappa\mid b^{(2)})\geq1-\alpha$.  This guarantee
is model-conditional; Appendix~\ref{app:formal_problem_details} gives the
predictive construction and calibration interpretation.

Finally, the task-specific deployment rule is
\begin{equation}
    \pi_\kappa^{\mathrm{dep}}
    =
    \begin{cases}
        \pi_\kappa, & L_\kappa\geq J_{\min},\\
        \pi_\kappa^{\mathrm{fallback}}, & L_\kappa<J_{\min}.
    \end{cases}
    \label{eq:deployment_rule}
\end{equation}
EG-MPT prepares $\{(\pi_\kappa,L_\kappa)\}_{\kappa\in\mathcal K}$ before task
reveal; thresholding occurs only at deployment.

For either information set $\mathcal F$, write
$R_{\mathrm{ctrl}}(\mathcal F):=\sum_iw_iR_i(\mathcal F)$.
Also let $V_{B_2}(\mathcal F)$ be the minimum expected task-readiness loss
over non-anticipating Stage~2 strategies with budget $B_2$.

\begin{proposition}[Fixed-budget task-readiness dominance]
\label{prop:end_to_end_value}
Assume the joint change model is calibrated.  With the same Stage~2 budget
and action set, exact Bayes strategies satisfy
\begin{equation}
    R_{\mathrm{ctrl}}(\mathcal F_1)
    \leq R_{\mathrm{ctrl}}(\mathcal F_0),
    \qquad
    V_{B_2}(\mathcal F_1)
    \leq V_{B_2}(\mathcal F_0).
    \label{eq:task_weighted_value}
\end{equation}
\end{proposition}

Propositions~\ref{prop:severity_orthogonality}--\ref{prop:end_to_end_value}
separate an architectural invariance from Bayes-value guarantees conditional
on model calibration; proofs and strictness conditions appear in
Appendix~\ref{app:theory}.

\section{Experiments}

\paragraph{Baselines.}
Bandit-QCD balances accumulated change evidence with exploration
\citep{gopalan2021bandit}; SEPT repeats a fixed task-agnostic probing schedule
\citep{yang2020sept}; ASID-FIM selects A-optimal identification trajectories
\citep{memmel2024asid}; OPAX maximizes task-agnostic information gain
\citep{sukhija2023opax}; and Task-OED weights identification by downstream
control value \citep{wagenmaker2023optimal}.  Bayes-Risk follows the
decision-theoretic comparison of experiments \citep{blackwell1953equivalent},
while Random samples uniformly.  All methods share the trajectory library,
calibration data, physical budget, recovery backend, certificate, and fallback;
exact adaptations are in Appendix~\ref{app:experimental_details}.

\paragraph{Protocol.}
Detection is measured on changed trials, while selective return scores the
recovered-or-fallback policy actually deployed after certification.  Results
are averaged within each seed and then across five seed-level means.  Full
settings and metric definitions appear in
Appendix~\ref{app:experimental_details}.

\begin{table*}[t]
\centering
\caption{Dormant-actuator readiness across six MuJoCo systems.}
\label{tab:exp01_readiness}
\setlength{\tabcolsep}{1.5pt}
\scriptsize
\resizebox{\textwidth}{!}{%
\begin{tabular}{|l|*{12}{c|}}
\hline
& \multicolumn{2}{|c|}{Ant}
& \multicolumn{2}{|c|}{HalfCheetah}
& \multicolumn{2}{|c|}{Hopper}
& \multicolumn{2}{|c|}{Humanoid}
& \multicolumn{2}{|c|}{Swimmer}
& \multicolumn{2}{|c|}{Walker2d} \\
\hline
Method & Detection & Return & Detection & Return
& Detection & Return & Detection & Return
& Detection & Return & Detection & Return \\
\hline
Random & $8.5\pm2.9$ & $0.9653\pm0.0015$ & $8.5\pm6.5$ & $0.9622\pm0.0016$ & $88.0\pm6.2$ & $0.8881\pm0.0022$ & $3.5\pm2.2$ & $0.9919\pm0.0017$ & $99.0\pm2.2$ & $0.9952\pm0.0050$ & $30.5\pm8.6$ & $0.9552\pm0.0002$ \\
Bayes-Risk & $29.5\pm8.2$ & $0.9710\pm0.0031$ & $33.0\pm9.3$ & $0.9688\pm0.0035$ & \underline{$98.5\pm2.2$} & $0.8919\pm0.0053$ & \underline{$24.0\pm4.2$} & $0.9923\pm0.0019$ & $99.0\pm2.2$ & $0.9952\pm0.0050$ & \underline{$77.0\pm5.4$} & $0.9602\pm0.0030$ \\
\hline
Bandit-QCD & $6.0\pm2.9$ & \underline{$0.9770\pm0.0005$} & $19.5\pm6.0$ & $0.9653\pm0.0036$ & $93.5\pm4.9$ & $0.8919\pm0.0053$ & $0.0\pm0.0$ & \underline{$0.9925\pm0.0019$} & $99.0\pm2.2$ & $0.9952\pm0.0050$ & $49.5\pm4.8$ & $0.9581\pm0.0016$ \\
SEPT & $6.0\pm5.8$ & $0.9661\pm0.0016$ & $6.0\pm4.5$ & $0.9621\pm0.0016$ & $88.5\pm3.8$ & $0.8848\pm0.0004$ & $1.5\pm1.4$ & $0.9915\pm0.0020$ & $99.0\pm2.2$ & $0.9952\pm0.0050$ & $28.0\pm3.3$ & $0.9552\pm0.0002$ \\
ASID-FIM & $19.0\pm3.8$ & $0.9700\pm0.0010$ & $32.5\pm7.3$ & $0.9670\pm0.0023$ & $93.0\pm3.7$ & $0.8913\pm0.0056$ & $7.0\pm4.1$ & $0.9925\pm0.0020$ & \underline{$99.5\pm1.1$} & $0.9964\pm0.0025$ & $29.0\pm6.5$ & $0.9575\pm0.0020$ \\
OPAX & $32.5\pm4.7$ & $0.9712\pm0.0009$ & $30.5\pm4.8$ & $0.9683\pm0.0032$ & $97.5\pm2.5$ & $0.8919\pm0.0053$ & $20.0\pm4.7$ & $0.9919\pm0.0022$ & $99.0\pm2.2$ & $0.9952\pm0.0050$ & \underline{$77.0\pm5.7$} & $0.9604\pm0.0029$ \\
Task-OED & $28.5\pm2.9$ & $0.9736\pm0.0013$ & $33.5\pm8.2$ & $0.9693\pm0.0027$ & $97.5\pm2.5$ & $0.8919\pm0.0053$ & $20.5\pm3.3$ & $0.9918\pm0.0015$ & $99.0\pm2.2$ & $0.9952\pm0.0050$ & $70.0\pm4.0$ & $0.9589\pm0.0031$ \\
\hline
\textbf{EG-MPT (Ours)} & \underline{$55.5\pm3.3$} & $0.9730\pm0.0011$ & \underline{$75.0\pm7.3$} & \underline{$0.9874\pm0.0036$} & \underline{$98.5\pm2.2$} & \underline{$0.9941\pm0.0021$} & $16.5\pm1.4$ & $0.9922\pm0.0018$ & \underline{$99.5\pm1.1$} & \underline{$0.9989\pm0.0025$} & $73.5\pm2.9$ & \underline{$0.9905\pm0.0014$} \\
\hline
\end{tabular}%
}
\end{table*}

\begin{table*}[t]
\centering
\caption{Readiness under recovery budget $B_2$ and pre-reveal probe limit $P$.}
\label{tab:exp04_05_budget_deadline}
\setlength{\tabcolsep}{1.5pt}
\scriptsize
\resizebox{\textwidth}{!}{%
\begin{tabular}{|l|c|c|c|c|c|c||c|c|c|c|c|c|}
\hline
& \multicolumn{6}{|c||}{Recovery Budget}
& \multicolumn{6}{c|}{Deadline Scarcity} \\
\hline
& \multicolumn{2}{|c|}{$B_2=0$}
& \multicolumn{2}{|c|}{$B_2=1$}
& \multicolumn{2}{|c||}{$B_2=2$}
& \multicolumn{2}{c|}{$P=2$}
& \multicolumn{2}{|c|}{$P=4$}
& \multicolumn{2}{|c|}{$P=6$} \\
\hline
Method & Regret $\downarrow$ & Return $\uparrow$
& Regret $\downarrow$ & Return $\uparrow$
& Regret $\downarrow$ & Return $\uparrow$
& Detection $\uparrow$ & Return $\uparrow$
& Detection $\uparrow$ & Return $\uparrow$
& Detection $\uparrow$ & Return $\uparrow$ \\
\hline
Random & $0.771\pm0.006$ & $0.769\pm0.002$
& $0.796\pm0.008$ & $0.769\pm0.002$
& $0.820\pm0.009$ & $0.772\pm0.003$
& $0.090\pm0.028$ & $0.804\pm0.009$
& $0.133\pm0.020$ & $0.803\pm0.009$
& $0.170\pm0.043$ & $0.805\pm0.008$ \\
Bayes-Risk & $0.765\pm0.006$ & $0.770\pm0.002$
& $0.804\pm0.008$ & $0.777\pm0.003$
& $0.837\pm0.012$ & $0.798\pm0.003$
& $0.150\pm0.023$ & $0.808\pm0.008$
& $0.365\pm0.011$ & $0.824\pm0.009$
& $0.496\pm0.009$ & $0.830\pm0.010$ \\
\hline
Bandit-QCD & $0.755\pm0.006$ & $0.769\pm0.002$
& $0.784\pm0.007$ & $0.771\pm0.001$
& $0.808\pm0.008$ & $0.786\pm0.006$
& $0.003\pm0.003$ & $0.804\pm0.008$
& $0.034\pm0.012$ & $0.805\pm0.009$
& $0.270\pm0.024$ & $0.821\pm0.010$ \\
SEPT & $0.769\pm0.006$ & $0.769\pm0.002$
& $0.792\pm0.006$ & $0.769\pm0.002$
& $0.815\pm0.007$ & $0.770\pm0.002$
& $0.049\pm0.030$ & $0.803\pm0.009$
& $0.100\pm0.015$ & $0.803\pm0.009$
& $0.152\pm0.017$ & $0.804\pm0.009$ \\
ASID-FIM & $0.764\pm0.006$ & $0.770\pm0.002$
& $0.792\pm0.007$ & $0.774\pm0.001$
& $0.817\pm0.010$ & $0.792\pm0.005$
& $0.100\pm0.035$ & $0.810\pm0.009$
& $0.163\pm0.023$ & $0.818\pm0.011$
& $0.291\pm0.026$ & $0.828\pm0.008$ \\
OPAX & $0.766\pm0.006$ & $0.770\pm0.002$
& $0.804\pm0.007$ & $0.778\pm0.003$
& $0.839\pm0.011$ & $0.799\pm0.003$
& $0.129\pm0.014$ & $0.812\pm0.009$
& $0.355\pm0.026$ & $0.825\pm0.007$
& $0.500\pm0.026$ & $0.831\pm0.008$ \\
Task-OED & $0.765\pm0.006$ & $0.770\pm0.002$
& $0.801\pm0.007$ & $0.778\pm0.002$
& $0.832\pm0.008$ & $0.796\pm0.003$
& $0.150\pm0.023$ & $0.808\pm0.008$
& $0.316\pm0.009$ & $0.820\pm0.008$
& $0.472\pm0.013$ & $0.829\pm0.010$ \\
\hline
\textbf{EG-MPT (Ours)} & \underline{$0.699\pm0.007$} & \underline{$0.906\pm0.007$}
& \underline{$0.741\pm0.007$} & \underline{$0.939\pm0.005$}
& \underline{$0.793\pm0.007$} & \underline{$0.940\pm0.005$}
& \underline{$0.198\pm0.020$} & \underline{$0.841\pm0.012$}
& \underline{$0.475\pm0.017$} & \underline{$0.894\pm0.011$}
& \underline{$0.689\pm0.014$} & \underline{$0.931\pm0.012$} \\
\hline
\end{tabular}
}
\end{table*}

\subsection{Readiness under Scarce Evidence}
\label{sec:experiments_mujoco}

\paragraph{Calibration sensitivity.}
Figure~\ref{fig:exp06_sensitivity} varies categorical resolution and labelled
calibration data.  Within-bin amplitude remains useful with a coarse channel,
whereas additional calibration mainly improves certificate reliability.

\noindent\begin{minipage}{\columnwidth}
\centering
\includegraphics[width=0.95\columnwidth]{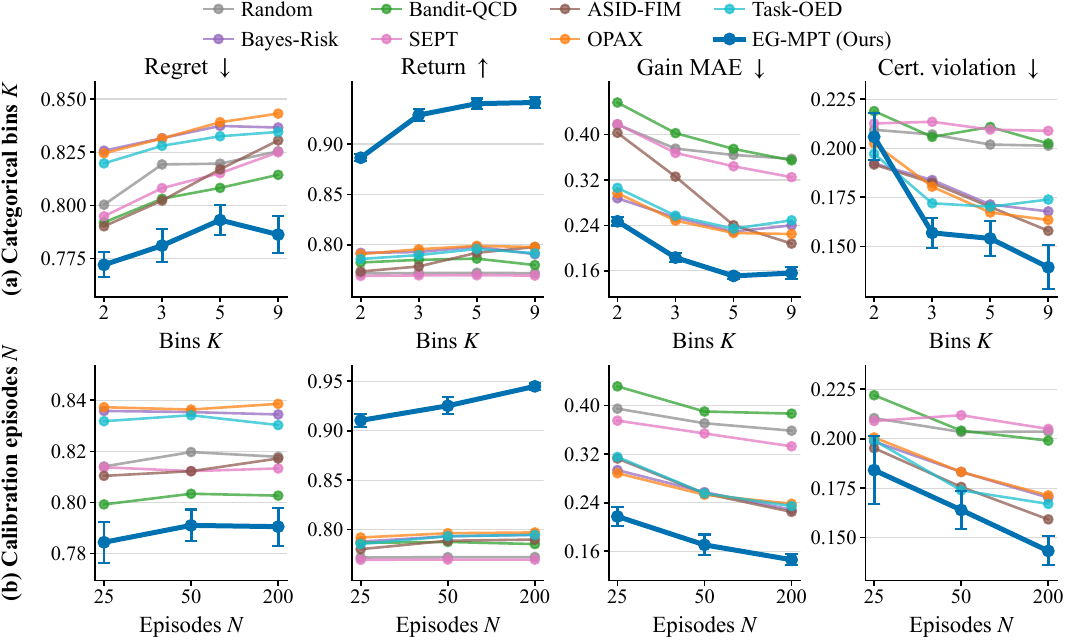}
\captionof{figure}{Sensitivity to categorical bins $K$ and calibration episodes
$N$ on 4 MuJoCo environments.}
\label{fig:exp06_sensitivity}
\end{minipage}

\paragraph{End-to-end readiness.}
Table~\ref{tab:exp01_readiness} covers 6 dormant-actuator MuJoCo systems.
EG-MPT gives the strongest selective return on four; detection and deployment
do not rank identically because evidence is valuable only when it supports an
effective correction.

\paragraph{Budgets and deadlines.}
Table~\ref{tab:exp04_05_budget_deadline} varies recovery interaction and probes
before task reveal.  EG-MPT leads even with zero recovery trajectories or only
two probing opportunities, where Stage~1 evidence cannot be replaced by
extensive data.

\subsection{Severity and Transfer}
\label{sec:experiments_transfer}

\paragraph{Severe faults.}
Table~\ref{tab:exp02_severe_faults} shows that the return advantage widens as
retained actuator effectiveness decreases, when estimating a continuous
correction matters most.

\begin{table}[H]
\centering
\caption{Severity across four MuJoCo environments.}
\label{tab:exp02_severe_faults}
\setlength{\tabcolsep}{2pt}
\scriptsize
\resizebox{\columnwidth}{!}{%
\begin{tabular}{|l|c|c|c|c|}
\hline
& \multicolumn{2}{|c|}{Severe fault ($g=0.20$)}
& \multicolumn{2}{|c|}{Calibrated fault ($g=0.35$)} \\
\hline
Method & Regret $\downarrow$ & Return $\uparrow$
& Regret $\downarrow$ & Return $\uparrow$ \\
\hline
Random & $0.854\pm0.008$ & $0.678\pm0.004$
& $0.820\pm0.009$ & $0.772\pm0.003$ \\
Bayes-Risk & $0.879\pm0.011$ & $0.689\pm0.003$
& $0.837\pm0.012$ & $0.798\pm0.003$ \\
\hline
Bandit-QCD & $0.847\pm0.008$ & $0.678\pm0.003$
& $0.808\pm0.008$ & $0.786\pm0.006$ \\
SEPT & $0.850\pm0.007$ & $0.676\pm0.002$
& $0.815\pm0.007$ & $0.770\pm0.002$ \\
ASID-FIM & $0.853\pm0.009$ & $0.684\pm0.003$
& $0.817\pm0.010$ & $0.792\pm0.005$ \\
OPAX & $0.880\pm0.011$ & $0.687\pm0.005$
& $0.839\pm0.011$ & $0.799\pm0.003$ \\
Task-OED & $0.876\pm0.009$ & $0.689\pm0.002$
& $0.832\pm0.008$ & $0.796\pm0.003$ \\
\hline
\textbf{EG-MPT (Ours)} & \underline{$0.819\pm0.008$} & \underline{$0.850\pm0.009$}
& \underline{$0.793\pm0.007$} & \underline{$0.940\pm0.005$} \\
\hline
\end{tabular}
}
\end{table}

\paragraph{Cross-physics transfer.}
Table~\ref{tab:exp10_cross_physics} moves from MuJoCo to BipedalWalker joint
torques and CarRacing steering/braking.  Transport leads for severe faults,
narrows at moderate loss, and disappears when the matched amplitude becomes
noise-dominated.

\begin{table}[!ht]
\centering
\caption{Cross-physics transfer on Gymnasium Box2D. A
nominal ensemble trained on one physics is deployed un-
der a dormant actuator fault in another: BipedalWalker and CarRacing. Entries are future-task regret, macro-averaged over the two backends.}
\label{tab:exp10_cross_physics}
\setlength{\tabcolsep}{3.0pt}
\scriptsize
\resizebox{\columnwidth}{!}{%
\begin{tabular}{|l|ccc|}
\hline
& \multicolumn{3}{|c|}{Future-task regret, $g_{\mathrm{true}}$} \\
\hline
Method & $g=0.30$ & $g=0.50$ & $g=0.70$ \\
\hline
Random & $0.0974\pm0.0080$ & $0.0463\pm0.0034$ & $0.0216\pm0.0029$ \\
Bayes-Risk & $0.0971\pm0.0090$ & $0.0405\pm0.0034$ & \underline{$0.0087\pm0.0017$} \\
\hline
Bandit-QCD & $0.0968\pm0.0068$ & $0.0489\pm0.0035$ & $0.0253\pm0.0027$ \\
SEPT & $0.0966\pm0.0086$ & $0.0484\pm0.0044$ & $0.0235\pm0.0013$ \\
ASID-FIM & $0.0963\pm0.0092$ & $0.0408\pm0.0037$ & $0.0130\pm0.0030$ \\
OPAX & $0.0960\pm0.0087$ & \underline{$0.0381\pm0.0047$} & $0.0094\pm0.0011$ \\
Task-OED & $0.0971\pm0.0090$ & $0.0405\pm0.0034$ & \underline{$0.0087\pm0.0017$} \\
\hline
\textbf{EG-MPT (Ours)} & \underline{$0.0888\pm0.0085$}
& $0.0396\pm0.0025$ & \underline{$0.0087\pm0.0017$} \\
\hline
\end{tabular}%
}
\end{table}

\paragraph{No recovery interaction.}
Table~\ref{tab:exp09v2_severity_transport} isolates the transported severity
belief on the native Pendulum and Acrobot tasks with no Stage~2 probe.  This tests whether
matched Stage~1 pulses improve estimation before additional recovery
interaction.  Lower gain error and CRPS show that their benefit is already
present at this stage.

\begin{table}[!ht]
\centering
\caption{Pendulum-Acrobot transport without recovery.}
\label{tab:exp09v2_severity_transport}
\setlength{\tabcolsep}{2.3pt}
\scriptsize
\resizebox{\columnwidth}{!}{%
\begin{tabular}{|l|cc|cc|}
\hline
& \multicolumn{2}{|c|}{$g_{\mathrm{true}}=0.35$}
& \multicolumn{2}{c|}{$g_{\mathrm{true}}=0.50$} \\
\hline
Method & Gain MAE & Severity CRPS & Gain MAE & Severity CRPS \\
\hline
Random & $0.274\pm0.006$ & $0.216\pm0.027$
& $0.147\pm0.011$ & $0.116\pm0.019$ \\
Bayes-Risk & $0.251\pm0.007$ & $0.198\pm0.026$
& $0.121\pm0.005$ & $0.095\pm0.013$ \\
\hline
Bandit-QCD & $0.528\pm0.014$ & $0.416\pm0.057$
& $0.388\pm0.019$ & $0.306\pm0.047$ \\
SEPT & $0.258\pm0.002$ & $0.203\pm0.025$
& $0.137\pm0.023$ & $0.109\pm0.027$ \\
ASID-FIM & $0.291\pm0.001$ & $0.229\pm0.028$
& $0.172\pm0.011$ & $0.136\pm0.023$ \\
OPAX & $0.251\pm0.006$ & $0.197\pm0.026$
& $0.130\pm0.004$ & $0.102\pm0.014$ \\
Task-OED & $0.293\pm0.014$ & $0.231\pm0.035$
& $0.147\pm0.015$ & $0.116\pm0.022$ \\
\hline
\textbf{EG-MPT (Ours)} & \underline{$0.223\pm0.010$} & \underline{$0.176\pm0.025$}
& \underline{$0.107\pm0.004$} & \underline{$0.084\pm0.012$} \\
\hline
\end{tabular}%
}
\end{table}

\subsection{Decision Quality}
\label{sec:experiments_decision_quality}

\paragraph{Evidence gate.}
Table~\ref{tab:expN7_transport_ablation} removes only the positive-evidence
gate.  The ungated variant deploys more often and slightly lowers raw regret,
but worsens severity error and certificate violations; the gate prevents
nominal-supporting fluctuations from becoming unsupported confidence.

\begin{table}[!ht]
\centering
\caption{Evidence-gate ablation on unseen-severity Box2D.}
\label{tab:expN7_transport_ablation}
\setlength{\tabcolsep}{2.3pt}
\scriptsize
\resizebox{\columnwidth}{!}{%
\begin{tabular}{|l|ccc|cc|}
\hline
& \multicolumn{3}{|c|}{Severity and return}
& \multicolumn{2}{c|}{Certification} \\
\hline
Method & Gain MAE & CRPS & Regret & Violation & Abstention \\
\hline
EG-MPT w/o gate & $0.2687$ & $0.2379$ & \underline{$0.0802$}
& $0.2927$ & $0.1529$ \\
& $\pm0.0263$ & $\pm0.0262$ & $\pm0.0023$ & $\pm0.0137$ & $\pm0.0174$ \\
\hline
\textbf{EG-MPT (Ours)} & \underline{$0.2225$} & \underline{$0.1969$} & $0.0834$
& \underline{$0.2672$} & $0.6373$ \\
& $\pm0.0155$ & $\pm0.0131$ & $\pm0.0021$ & $\pm0.0209$ & $\pm0.0406$ \\
\hline
\end{tabular}%
}
\end{table}

\paragraph{Mechanism diagnostics.}
At the shared recovery budget $B_2=2$,
Table~\ref{tab:exp04_diagnostics} separates the intermediate
decisions from end-to-end utility.  EG-MPT has the lowest gain error, highest
detection rate, lowest certificate violation rate, and shortest detection
delay.  These results match the intended mechanism:
the evidence gate protects localization while the retained matched coordinate
supplies severity information to recovery and certification.

\begin{table}[!ht]
\centering
\caption{Mechanism-level diagnostics at $B_2=2$.}
\label{tab:exp04_diagnostics}
\setlength{\tabcolsep}{2.8pt}
\scriptsize
\resizebox{\columnwidth}{!}{%
\begin{tabular}{|l|c|c|c|c|}
\hline
Method & Gain MAE $\downarrow$ & Detection $\uparrow$
& Cert. violation $\downarrow$ & Detection delay $\downarrow$ \\
\hline
Random & $0.364\pm0.018$ & $0.339\pm0.045$ & $0.202\pm0.012$ & $20.280\pm0.381$ \\
Bayes-Risk & $0.229\pm0.017$ & $0.595\pm0.043$ & $0.171\pm0.019$ & $17.055\pm0.473$ \\
Bandit-QCD & $0.375\pm0.016$ & $0.421\pm0.027$ & $0.211\pm0.006$ & $19.707\pm0.402$ \\
SEPT & $0.344\pm0.005$ & $0.321\pm0.027$ & $0.210\pm0.010$ & $21.390\pm0.278$ \\
ASID-FIM & $0.240\pm0.012$ & $0.434\pm0.020$ & $0.170\pm0.007$ & $20.475\pm0.241$ \\
OPAX & $0.227\pm0.018$ & $0.594\pm0.034$ & $0.167\pm0.012$ & $16.981\pm0.294$ \\
Task-OED & $0.234\pm0.013$ & $0.574\pm0.021$ & $0.170\pm0.011$ & $17.367\pm0.326$ \\
\hline
\textbf{EG-MPT} & \underline{$0.151\pm0.005$} & \underline{$0.756\pm0.019$}
& \underline{$0.154\pm0.009$} & \underline{$14.894\pm0.272$} \\
\hline
\end{tabular}
}
\end{table}

\begin{table*}[t]
\centering
\caption{Future-task risk profiles on Ant, HalfCheetah, and Walker2d.}
\label{tab:exp07_risk_profiles}
\setlength{\tabcolsep}{1.4pt}
\scriptsize
\resizebox{\textwidth}{!}{%
\begin{tabular}{|l|c|c|c||c|c|c||c|c|c|}
\hline
& \multicolumn{3}{|c||}{Uniform}
& \multicolumn{3}{c||}{Head-heavy}
& \multicolumn{3}{c|}{Rare-critical} \\
\hline
Method & Return $\uparrow$ & Gain MAE $\downarrow$ & Cert. viol. $\downarrow$
& Return $\uparrow$ & Gain MAE $\downarrow$ & Cert. viol. $\downarrow$
& Return $\uparrow$ & Gain MAE $\downarrow$ & Cert. viol. $\downarrow$ \\
\hline
Random & $0.789\pm0.003$ & $0.461\pm0.019$ & $0.265\pm0.016$
& $0.789\pm0.003$ & $0.461\pm0.019$ & $0.258\pm0.016$
& $0.789\pm0.003$ & $0.461\pm0.019$ & $0.258\pm0.016$ \\
Bayes-Risk & $0.819\pm0.010$ & $0.306\pm0.028$ & $0.225\pm0.020$
& $0.816\pm0.008$ & $0.332\pm0.011$ & $0.228\pm0.012$
& $0.816\pm0.008$ & $0.334\pm0.014$ & $0.225\pm0.012$ \\
\hline
Bandit-QCD & $0.805\pm0.009$ & $0.486\pm0.017$ & $0.277\pm0.008$
& $0.805\pm0.009$ & $0.486\pm0.017$ & $0.272\pm0.012$
& $0.805\pm0.009$ & $0.486\pm0.017$ & $0.272\pm0.012$ \\
SEPT & $0.789\pm0.002$ & $0.447\pm0.004$ & $0.272\pm0.014$
& $0.789\pm0.002$ & $0.447\pm0.004$ & $0.261\pm0.015$
& $0.789\pm0.002$ & $0.447\pm0.004$ & $0.261\pm0.015$ \\
ASID-FIM & $0.813\pm0.007$ & $0.313\pm0.016$ & $0.217\pm0.011$
& $0.813\pm0.007$ & $0.313\pm0.016$ & $0.202\pm0.013$
& $0.813\pm0.007$ & $0.313\pm0.016$ & $0.202\pm0.013$ \\
OPAX & $0.822\pm0.007$ & $0.298\pm0.023$ & $0.220\pm0.015$
& $0.822\pm0.007$ & $0.298\pm0.023$ & $0.215\pm0.015$
& $0.822\pm0.007$ & $0.298\pm0.023$ & $0.215\pm0.015$ \\
Task-OED & $0.821\pm0.008$ & $0.308\pm0.021$ & $0.223\pm0.017$
& $0.815\pm0.003$ & $0.349\pm0.011$ & $0.238\pm0.016$
& $0.812\pm0.003$ & $0.351\pm0.019$ & $0.235\pm0.013$ \\
\hline
\textbf{EG-MPT (Ours)} & \underline{$0.927\pm0.005$} & \underline{$0.166\pm0.010$} & \underline{$0.170\pm0.010$}
& \underline{$0.918\pm0.012$} & \underline{$0.214\pm0.020$} & \underline{$0.185\pm0.022$}
& \underline{$0.914\pm0.010$} & \underline{$0.206\pm0.024$} & \underline{$0.179\pm0.020$} \\
\hline
\end{tabular}%
}
\end{table*}

\subsection{Breadth and Risk}

\paragraph{Compact plants.}
Table~\ref{tab:expM1_estimation} evaluates policy-consumed effectiveness
estimates across aircraft, attitude control, servos, quadrotors, buildings, and
turbines.  EG-MPT leads both metrics on the six shown families; the complete
eleven-family matrix is in Appendix~\ref{app:experimental_details}.

\begin{table}[!ht]
\centering
\caption{Effectiveness estimates on six compact-plant families; the
full matrix is in Appendix~\ref{tab:expM1_full_matrix}.}
\label{tab:expM1_estimation}
\setlength{\tabcolsep}{1.8pt}
\scriptsize
\resizebox{\columnwidth}{!}{%
\begin{tabular}{|l|cccccc|c|}
\hline
Method & Aircraft & Attitude & Servo & Quadrotor & Building & Turbine & Macro \\
\hline
\multicolumn{8}{|l|}{\textit{Compensation error} $|c_j g_j - 1|$} \\
\hline
Random & $0.490$ & $0.497$ & $0.490$ & $0.530$ & $0.528$ & $0.478$ & $0.502$ \\
Bayes-Risk & $0.283$ & $0.283$ & $0.296$ & $0.399$ & $0.389$ & $0.291$ & $0.324$ \\
Bandit-QCD & $0.486$ & $0.486$ & $0.485$ & $0.540$ & $0.545$ & $0.494$ & $0.506$ \\
SEPT & $0.474$ & $0.468$ & $0.461$ & $0.500$ & $0.494$ & $0.474$ & $0.479$ \\
ASID-FIM & $0.286$ & $0.283$ & $0.292$ & $0.395$ & $0.399$ & $0.304$ & $0.327$ \\
OPAX & $0.318$ & $0.321$ & $0.329$ & $0.420$ & $0.429$ & $0.306$ & $0.354$ \\
Task-OED & $0.329$ & $0.329$ & $0.339$ & $0.431$ & $0.437$ & $0.309$ & $0.362$ \\
\textbf{EG-MPT (Ours)} & \underline{$0.268$} & \underline{$0.264$}
& \underline{$0.280$} & \underline{$0.377$} & \underline{$0.381$}
& \underline{$0.284$} & \underline{$0.309$} \\
\hline
\multicolumn{8}{|l|}{\textit{Worst-mechanism effectiveness error}} \\
\hline
Random & $0.288$ & $0.293$ & $0.286$ & $0.301$ & $0.299$ & $0.270$ & $0.289$ \\
Bayes-Risk & $0.172$ & $0.172$ & $0.180$ & $0.213$ & $0.205$ & $0.166$ & $0.185$ \\
Bandit-QCD & $0.364$ & $0.363$ & $0.363$ & $0.377$ & $0.379$ & $0.362$ & $0.368$ \\
SEPT & $0.263$ & $0.255$ & $0.252$ & $0.262$ & $0.263$ & $0.257$ & $0.259$ \\
ASID-FIM & $0.176$ & $0.173$ & $0.183$ & $0.207$ & $0.211$ & $0.177$ & $0.188$ \\
OPAX & $0.198$ & $0.201$ & $0.206$ & $0.231$ & $0.228$ & $0.175$ & $0.206$ \\
Task-OED & $0.204$ & $0.206$ & $0.214$ & $0.239$ & $0.235$ & $0.177$ & $0.212$ \\
\textbf{EG-MPT (Ours)} & \underline{$0.172$} & \underline{$0.172$}
& \underline{$0.180$} & \underline{$0.207$} & \underline{$0.203$}
& \underline{$0.162$} & \underline{$0.183$} \\
\hline
\end{tabular}%
}
\end{table}

\paragraph{Unseen severity.}
Table~\ref{tab:expN7_unseen_severity} tests gains absent from the categorical
calibration channel.  Transport remains useful for strong loss, but converges
to the comparison methods as mild signatures approach the noise floor.

\begin{table}[!ht]
\centering
\caption{Box2D readiness at gains absent from categorical calibration.}
\label{tab:expN7_unseen_severity}
\setlength{\tabcolsep}{3.0pt}
\scriptsize
\resizebox{\columnwidth}{!}{%
\begin{tabular}{|l|ccc|}
\hline
& \multicolumn{3}{|c|}{Future-task regret, unseen $g_{\mathrm{true}}$} \\
\hline
Method & $g=0.20$ & $g=0.50$ & $g=0.65$ \\
\hline
Random & $0.1381\pm0.0090$ & $0.0812\pm0.0039$ & $0.0643\pm0.0047$ \\
Bayes-Risk & $0.1326\pm0.0072$ & $0.0768\pm0.0040$ & $0.0567\pm0.0042$ \\
\hline
Bandit-QCD & $0.1359\pm0.0090$ & $0.0802\pm0.0037$ & $0.0709\pm0.0042$ \\
SEPT & $0.1318\pm0.0058$ & $0.0860\pm0.0064$ & $0.0714\pm0.0091$ \\
ASID-FIM & $0.1318\pm0.0051$ & $0.0722\pm0.0093$ & $0.0584\pm0.0022$ \\
OPAX & $0.1336\pm0.0053$ & \underline{$0.0683\pm0.0057$} & \underline{$0.0523\pm0.0041$} \\
Task-OED & $0.1326\pm0.0072$ & $0.0768\pm0.0040$ & $0.0567\pm0.0042$ \\
\hline
\textbf{EG-MPT (Ours)} & \underline{$0.1230\pm0.0067$}
& $0.0705\pm0.0038$ & $0.0567\pm0.0041$ \\
\hline
\end{tabular}%
}
\end{table}

\paragraph{Future-task risk.}
Table~\ref{tab:exp07_risk_profiles} changes the task distribution from uniform
to head-heavy and rare-critical profiles.  The consistent advantage shows that
task-agnostic preparation must still account for downstream consequence.

\paragraph{Synthesis.}
The benefit is not simply higher change-detection frequency.  Readiness
improves when evidence answers two control questions: which correction is
supported and whether its uncertainty justifies deployment.  The zero-probe
and severe-fault studies isolate the first role; risk profiles and the gate
ablation isolate the second.  Cross-physics and compact-plant results show that
this evidence reuse is not tied to one simulator.  Conversely, convergence on
mild faults identifies the boundary: when matched responses approach the noise
floor, transport adds little.  EG-MPT therefore saves scarce recovery
interaction rather than replacing recovery data or calibration.

Selective return and certificate violations expose complementary failure
modes.  Aggressive deployment can reduce raw regret while weakening
reliability, whereas excessive abstention can hide usable capability.
Reporting both makes the fallback cost explicit and prevents a method from
appearing safe merely by deferring every difficult task.
These metrics also rule out different shortcuts.  Detection alone cannot tell
whether an alert supports the correct compensation; return alone can reward
aggressive deployment despite unreliable certificates; and violations alone
can be minimized by always abstaining.  Reporting gain error, selective return,
violations, abstention, delay, and interaction cost therefore evaluates the
complete readiness decision rather than one favorable component.  This is why
method rankings change across diagnosis, recovery, and deployment.
\section{Conclusion}

EG-MPT unifies active localization, severity-aware recovery, and task-wise
certification under a task-agnostic budget.  Across MuJoCo, Box2D, Classic
Control, and compact plants, it is most useful when recovery evidence is scarce
and actuator loss matters.  Evidence gating limits unsupported confidence,
while mild faults reveal the noise boundary.  Extending calibrated transport
beyond localized effectiveness changes is the main challenge.

\bibliography{ref1}

\newpage

\onecolumn
\setlength{\oddsidemargin}{0.25in}
\setlength{\evensidemargin}{0.25in}
\setlength{\textwidth}{6.0in}
\setlength{\topmargin}{0.25in}
\setlength{\textheight}{8.0in}
\setlength{\columnwidth}{\textwidth}
\setlength{\linewidth}{\textwidth}
\hsize=\textwidth
\vsize=\textheight
\makeatletter
\setlength{\@colroom}{\textheight}
\setlength{\@colht}{\textheight}
\makeatother
\appendix
\setcounter{secnumdepth}{2}

\hrule
\begin{center}
  {\LARGE\bfseries Appendix\par}
  \vspace{0.4em}
  {\Large\itshape Diagnose, Recover, Certify:\\Task Readiness under Hidden Dynamics Changes\par}
\end{center}
\hrule

\vspace{1em}
\makeatletter
\renewcommand{\section}{\@startsection{section}{1}{\z@}%
  {-2.0ex plus -0.5ex minus -.2ex}%
  {3pt plus 2pt minus 1pt}%
  {\Large\bf\raggedright}}
\makeatother
\renewcommand{\thesection}{\Alph{section}}
\renewcommand{\thesubsection}{\thesection.\arabic{subsection}}
\renewcommand{\labelenumi}{\arabic{enumi}.}

\makeatletter
\newcommand{\appendixcontents}{%
  {\large\bfseries Contents\par}
  \vspace{0.4em}
  \@starttoc{atoc}}
\newcommand{\appsection}[1]{%
  \section{#1}%
  \addcontentsline{atoc}{section}{\protect\numberline{\thesection}#1}}
\newcommand{\appsubsection}[1]{%
  \subsection{#1}%
  \addcontentsline{atoc}{subsection}{\protect\numberline{\thesubsection}#1}}
\makeatother

\appendixcontents

\newpage
\appsection{Related Work}
\label{app:related_work}

\appsubsection{Non-Stationary Reinforcement Learning}

Non-stationary RL typically models an environment whose reward or transition
kernel changes over time. FANS-RL represents this change through localized
latent factors and causal structure \citep{feng2022fans}; continual adaptation
learns a latent representation from current and past experience
\citep{xie2021continual}; and regret-oriented methods analyze piecewise-changing
MDPs \citep{chen2022goaloriented}. Detection-and-adaptation methods also handle
unknown change points by monitoring policy-behavior shifts and regularizing
post-change adaptation \citep{liu2024bada}. These formulations primarily ask
how to continue improving return while the environment evolves, often treating
a shift as visible once it affects recent reward or transition statistics.

Changepoint methods instead focus on controlling false alarms and detection
delay. R-BOCPD-UCRL2 combines restarted Bayesian online changepoint detection
with optimistic RL in piecewise-stationary MDPs \citep{alami2023rbocpd}, while
robust stationarity tests detect shifts from offline sequential-decision data
while controlling type-I error \citep{wang2023stationarity}. Active
quickest-change methods select costly sensing channels to balance detection
delay, exploration, and false alarms \citep{gopalan2021bandit}. Their primary
objective remains detection. The dormant-drift setting considered here is
stricter: the nominal and changed systems can induce identical trajectory laws
under the deployed policy. This creates an information-theoretic failure of
passive detection, not merely a low-power statistical test, and motivates
interventions whose evidence also supports capability recovery.

\appsubsection{Causal Reinforcement Learning}

In causal and control settings, a system may be identifiable only after the
agent perturbs variables that distinguish competing mechanisms. The
``Learning by Doing'' benchmark explicitly connects excitation in control,
causal identifiability, and reinforcement learning by providing observational
and interventional dynamical data \citep{weichwald2022learning}. Related
identifiability results show how sufficiently diverse interventions can recover
system parameters in linear time-invariant systems that remain ambiguous from
passive observations \citep{rajendran2024interventional}.

\paragraph{Dormant faults.}
Our formulation makes this distinction operational: a mechanism can be
physically degraded yet statistically indistinguishable from nominal behavior
because the active policy does not excite it. Unlike generic causal discovery,
we do not attempt to reconstruct a complete causal graph. Instead, Stage~1
uses a calibrated collection of mechanism-targeted trajectories to isolate the
location of a local change, while preserving an explicit separation between
localization evidence and severity inference.

\appsubsection{Active System Identification}

\paragraph{Experiment design.}
Active identification selects experiments that reduce uncertainty about
dynamics before downstream control. SEPT learns a probing behavior that infers
latent dynamics before executing a universal policy \citep{yang2020sept}; OPAX
explores uncertain dynamics for zero-shot planning across multiple downstream
tasks \citep{sukhija2023opax}; and ASID uses a simulator to design informative
real-world exploration for physical-parameter identification
\citep{memmel2024asid}. These methods establish that interaction data should
be selected rather than collected indiscriminately.

Optimal experiment design makes the connection to control explicit by weighting
parameters according to their downstream controller loss
\citep{wagenmaker2023optimal}. EG-MPT similarly assigns scarce trajectories by
their decision value, but its budget comes after an alarm and before the future
task is revealed. Consequently, it must retain information useful for a
distribution of possible tasks, rather than tune a probe to one known terminal
objective or estimate every parameter to uniform accuracy.

\appsubsection{Meta-Reinforcement Learning}

Meta-RL and Bayes-adaptive methods infer latent context from a short history and
condition a policy on that context. PEARL represents task uncertainty with a
probabilistic context variable and adapts off policy \citep{rakelly2019pearl};
SEPT likewise converts a probe trajectory into a latent dynamics estimate
before deployment \citep{yang2020sept}. This literature is especially relevant
because it treats uncertainty as a control input rather than only a prediction
error.

\paragraph{Task-agnostic preparation.}
In standard rapid adaptation, the task is usually fixed or its distribution is
available during data collection. Our recovery period is deliberately
task-agnostic: observations must be selected before the identity of the next
task is known. The output is therefore not one adapted policy, but a posterior
that can be propagated into a family of task-conditioned recovery policies and
readiness decisions after the task is revealed.

\appsubsection{Robust Reinforcement Learning}

Off-dynamics RL asks whether a policy or value estimate learned in one dynamics
can be used in another. ODRL provides a benchmark spanning online and offline
transfer \citep{lyu2024odrl}; distributionally robust off-dynamics RL protects
against a set of target kernels \citep{liu2024offdynamics}; and model-based
approaches learn target-domain rollouts from limited target data
\citep{guo2026mobody}. These works emphasize distribution shift between source
and target environments rather than the detection of the shift itself.

\paragraph{Localized uncertainty.}
The relevant uncertainty in our setting is localized to a changed mechanism and
its effectiveness, with a nominal model still useful elsewhere. This differs
from a global ambiguity set over transitions: localization determines which
correction and which recovery trajectories are useful. The evidence-gated
posterior consequently separates uncertainty that affects a task's attainable
return from uncertainty that has no effect on its required mechanisms.

\appsubsection{Selective Decision-Making}

Safe policy improvement avoids unsupported updates by reverting toward a
baseline \citep{laroche2019spibb}; safe model-based RL derives stability
guarantees while learning and expanding a safe region
\citep{berkenkamp2017safe}; and sampling-based safe RL enforces constraints
throughout learning and deployment \citep{suttle2024sampling}. These approaches
address constraints or conservative improvement during online control, which is
complementary to deciding whether a recovered capability is adequate for a
future task.

\paragraph{Readiness and abstention.}
Selective prediction formalizes a coverage--risk trade-off by allowing a model
to reject uncertain inputs; SelectiveNet jointly optimizes prediction and a
reject option \citep{geifman2019selectivenet}. EG-MPT uses the same decision
principle at the control level: for each revealed task, it compares a
posterior-predictive lower confidence bound with a required threshold and
either deploys a recovered policy or abstains to a fallback. The certificate is
therefore task-wise, tied to a finite recovery budget, and explicitly informed
by both fault and model uncertainty.

Table~\ref{tab:related_work_scope} summarizes the combination of capabilities
needed here. Existing families cover valuable pieces---change detection,
intervention, task transfer, robust control, or abstention---but do not jointly
address a policy-hidden component change, task-agnostic recovery, and a
task-wise readiness certificate. The table describes the central emphasis of
each family rather than claiming that every individual paper has exactly the
same scope.

\begin{table}[h]
\centering
\caption{Scope of related-work families. $\checkmark$ denotes an explicitly
supported property, $\triangle$ a partial or indirect treatment, and -- a
property that is not central. \textsc{Dormant} denotes policy-induced zero
passive evidence; \textsc{Hidden task} means task identity is unknown during
adaptation; and \textsc{Certificate} denotes a task-wise certificate or
abstention.}
\label{tab:related_work_scope}
\resizebox{\textwidth}{!}{%
\begin{tabular}{|l|c|c|c|c|c|c|c|}
\hline
& \textsc{Change} & \textsc{Active} & \textsc{Dormant}
& \textsc{Hidden task} & \textsc{Task set} & \textsc{Certificate}
& \textsc{End-to-end} \\
\hline
\shortstack[l]{Non-stationary and adaptive RL\\[-0.2ex]
\citep{feng2022fans,xie2021continual,alami2023rbocpd,wang2023stationarity}}
& $\checkmark$ & $\triangle$ & -- & -- & $\triangle$ & $\triangle$ & -- \\[0.4ex]
\shortstack[l]{Causal control and fault diagnosis\\[-0.2ex]
\citep{weichwald2022learning,rajendran2024interventional}}
& $\triangle$ & $\checkmark$ & $\triangle$ & -- & -- & -- & -- \\[0.4ex]
\shortstack[l]{System identification and exploration\\[-0.2ex]
\citep{yang2020sept,sukhija2023opax,wagenmaker2023optimal,memmel2024asid}}
& $\triangle$ & $\checkmark$ & -- & $\triangle$ & $\checkmark$ & -- & -- \\[0.4ex]
\shortstack[l]{Meta-RL and transfer\\[-0.2ex]
\citep{rakelly2019pearl,lyu2024odrl,liu2024offdynamics,guo2026mobody}}
& $\triangle$ & $\triangle$ & -- & $\triangle$ & $\triangle$ & $\triangle$ & -- \\[0.4ex]
\shortstack[l]{Safe and selective decision-making\\[-0.2ex]
\citep{laroche2019spibb,berkenkamp2017safe,suttle2024sampling,geifman2019selectivenet}}
& $\triangle$ & $\triangle$ & -- & -- & -- & $\triangle$ & -- \\[0.4ex]
\hline
\textbf{Ours} & $\checkmark$ & $\checkmark$ & $\checkmark$
& $\checkmark$ & $\checkmark$ & $\checkmark$ & $\checkmark$ \\
\hline
\end{tabular}%
}
\end{table}

\newpage
\appsection{Formal Details of the Problem Formulation}
\label{app:formal_problem_details}

This appendix fixes the notation and probability objects used by the problem
formulation and EG-MPT.  It follows the indexing convention of the main text:
$i,k\in\{1,\ldots,m\}$ index mechanisms, $t$ indexes Stage~1 rounds,
$\ell$ indexes Stage~2 recovery rounds, and $q$ denotes a generic trajectory
in the library $\mathcal Q$.  The trajectories selected at the two stages are
therefore written $q_t$ and $q_\ell$, respectively; the star used in the
argmax definitions of the main text is suppressed after a maximizer has been
selected.  A Stage~1 diagnostic response is the stacked vector
$\mathbf x_{q_t}$ used to compute
$S_{q_t},C_{q_t},z_{q_t}$, whereas a Stage~2 recovery response is
$Y_\ell$.  Thus $q$ labels an intervention design, while $t$ and $\ell$ label
when an observation is collected.  In the controlled-dynamics equations,
$t$ retains its standard local meaning as an environment time step; in the
Stage~1 belief recursion it denotes a diagnostic round.  A diagnostic round
contains the multiple environment steps of its selected trajectory, so the
two uses are distinct.

The section proceeds in the same order as the method:
\begin{itemize}
    \item define the latent change and the sense in which it is dormant;
    \item show how Stage~1 interventions produce localization and transported
    severity evidence, then how $Y_\ell$ updates the Stage~2 joint posterior;
    and
    \item map the terminal posterior to recovered policies and readiness
    certificates.
\end{itemize}

\appsubsection{Change Model and Dormancy}

\paragraph{Piecewise dynamics.}
The latent dynamics state is $(H,\mathbf g)$.  Here
$H\in\mathcal H=\{0,1,\ldots,m\}$ is the change location:
$H=0$ denotes nominal dynamics, while $H=i>0$ denotes a change in mechanism
$i$.  The vector $\mathbf g=(g_1,\ldots,g_m)$ contains mechanism
effectiveness values.  Under $H=i>0$, only $g_i$ may differ from one; under
$H=0$, $\mathbf g=\mathbf1$.  Conditional on a possible change time $\tau$,
the controlled process evolves according to
\begin{equation}
    s_{t+1}\sim
    \begin{cases}
        P_{0,\mathbf{1}}(\cdot\mid s_t,a_t), & t<\tau,\\
        P_{H,\mathbf{g}}(\cdot\mid s_t,a_t), & t\geq\tau.
    \end{cases}
    \label{eq:appendix_piecewise_dynamics}
\end{equation}
For $H=0$, the second branch equals the first, so $\tau$ has no operational
effect.  A mechanism may represent an actuator, sensor, communication channel,
or any localized subsystem whose effectiveness changes the transition law.

\paragraph{Trajectory law.}
For a Markov policy $\pi$, define the post-change segment
\begin{equation}
    \xi_{\tau:T}
    :=
    (s_\tau,a_\tau,s_{\tau+1},a_{\tau+1},\ldots,s_T).
    \label{eq:appendix_trajectory_segment}
\end{equation}
Because the state distribution at $\tau$ is generated by the common nominal
process, it is the same under the nominal and changed hypotheses.  Conditional
on $s_\tau$, the law of the remaining segment factorizes as
\begin{equation}
    \mathbb{P}^{\pi}_{H,\mathbf{g},T}
    \bigl(\xi_{\tau:T}\mid s_\tau\bigr)
    =
    \prod_{t=\tau}^{T-1}
    \pi(a_t\mid s_t)\,
    P_{H,\mathbf{g}}(s_{t+1}\mid s_t,a_t).
    \label{eq:appendix_trajectory_law}
\end{equation}
For continuous spaces, the factors in
Eq.~\eqref{eq:appendix_trajectory_law} are densities with respect to a common
dominating measure.

\paragraph{A sufficient condition for dormancy.}
For a candidate change $(i,\mathbf g)$, define its change-sensitive
state--action set
\begin{equation}
    \mathcal X_i(\mathbf g)
    :=
    \left\{
        (s,a):
        P_{i,\mathbf g}(\cdot\mid s,a)
        \neq
        P_{0,\mathbf 1}(\cdot\mid s,a)
    \right\}.
    \label{eq:appendix_change_sensitive_set}
\end{equation}
If the deployed policy has zero occupancy on this set,
\begin{equation}
    \mathbb P^{\pi_0}_{0,\mathbf1,R-1}
    \bigl((s_t,a_t)\in\mathcal X_i(\mathbf g)\bigr)
    =0,
    \qquad
    t=\tau,\ldots,R-1,
    \label{eq:appendix_zero_sensitive_occupancy}
\end{equation}
then the two transition kernels agree almost surely at every state--action pair
encountered by $\pi_0$.  Induction through the product in
Eq.~\eqref{eq:appendix_trajectory_law} gives
$\mathbb P^{\pi_0}_{i,\mathbf g,T}
=\mathbb P^{\pi_0}_{0,\mathbf1,T}$ for every
$T\in\{\tau,\ldots,R-1\}$.  Consequently, for every bounded measurable
statistic $f$ of the passive trajectory,
\begin{equation}
    \mathbb E_{\xi\sim
        \mathbb P^{\pi_0}_{i,\mathbf g,T}}
    [f(\xi)]
    =
    \mathbb E_{\xi\sim
        \mathbb P^{\pi_0}_{0,\mathbf1,T}}
    [f(\xi)].
    \label{eq:appendix_passive_statistic_equivalence}
\end{equation}
Thus no passive statistic---not only the reward or a one-step prediction
error---can distinguish the two hypotheses.  In particular, their
Kullback--Leibler divergence is zero for every pre-reveal horizon.

\paragraph{Actuator-gain example.}
Suppose mechanism $i$ scales action coordinate $i$, and define
$a^{(i,g_i)}:=(a_1,\ldots,g_i a_i,\ldots,a_m)$.  The changed dynamics take the
form
\begin{equation}
    P_{i,\mathbf g}(\cdot\mid s,a)
    =
    P_{0,\mathbf1}(\cdot\mid s,a^{(i,g_i)}).
    \label{eq:appendix_gain_example}
\end{equation}
If the deployed policy never activates that coordinate,
\begin{equation}
    \pi_0\bigl(\{a\in\mathcal A:a_i=0\}\mid s\bigr)=1
    \quad
    \text{for every state in the support of its occupancy},
    \label{eq:appendix_zero_occupancy}
\end{equation}
then $a^{(i,g_i)}=a$ almost surely and therefore
\begin{equation}
    P_{i,\mathbf g}(\cdot\mid s,a)
    =
    P_{0,\mathbf1}(\cdot\mid s,a)
    \quad
    \pi_0\text{-almost surely}.
    \label{eq:appendix_kernel_agreement}
\end{equation}
This verifies Eq.~\eqref{eq:appendix_zero_sensitive_occupancy}.  A future task
may instead require $a_i\neq0$, so the same degradation can be irrelevant to
the present trajectory law yet consequential for future return.

\appsubsection{Interventions and Two-Stage Evidence}

\paragraph{Stage~1 diagnostic response.}
A diagnostic trajectory $q\in\mathcal Q$ is a finite open-loop sequence or
feedback controller with horizon $L_q$ and designated target $j(q)$.  Starting
from its prescribed reset distribution, it produces the stacked observed-state
response
\begin{equation}
    \mathbf x_q
    :=
    \operatorname{vec}(s_0,s_1,\ldots,s_{L_q}).
    \label{eq:appendix_diagnostic_response}
\end{equation}
The prescribed actions are part of $q$ and hence need not be repeated in the
random response.  Let $\mathbb P_q(\cdot\mid h,\mathbf g)$ denote the law of
$\mathbf x_q$ under
$(H,\mathbf g)=(h,\mathbf g)$.  Unlike passive operation, $q$ is designed to
visit the change-sensitive set.  It is informative for $(i,\mathbf g)$ whenever
\begin{equation}
    D_{\mathrm{KL}}
    \left(
        \mathbb P_q(\cdot\mid i,\mathbf g)
        \,\Vert\,
        \mathbb P_q(\cdot\mid0,\mathbf1)
    \right)
    >0.
    \label{eq:appendix_interventional_information}
\end{equation}
At Stage~1 round $t$, EG-MPT executes $q_t$, observes
$\mathbf x_{q_t}$, and constructs the residual
$\mathbf r_{q_t}$ and signature $\mathbf h_{q_t}$ from
Eq.~\eqref{eq:residual_signature}.  The same response yields the matched score
$S_{q_t}$, categorical outcome $C_{q_t}$, and normalized coordinate
$z_{q_t}$.  Only $C_{q_t}$ updates the location posterior $b_t$; when the
gate in Eq.~\eqref{eq:evidence_gate} opens, $z_{q_t}$ additionally updates
the conditional effectiveness posterior
$\varphi_{t,j(q_t)}$.  This is the precise separation between localization
and transported severity evidence used in the main methodology.

\paragraph{Finite recovery hypothesis space.}
For the gain grid $\mathcal G$ used by EG-MPT, define
\begin{equation}
    \Theta_{\mathcal G}
    :=
    \{(0,\mathbf1)\}
    \cup
    \bigcup_{i=1}^{m}
    \left\{
        (i,\mathbf g):
        g_i\in\mathcal G,\;
        g_k=1\ \forall k\neq i
    \right\}.
    \label{eq:appendix_recovery_hypothesis_space}
\end{equation}
Writing $\theta=(h,\mathbf g)\in\Theta_{\mathcal G}$,
Eq.~\eqref{eq:recovery_initialization} gives
$b_0^{(2)}(\theta)=b^{(1)}(h)p_0(\mathbf g\mid h)$.  A continuous gain model is
obtained by replacing sums over $\Theta_{\mathcal G}$ with the corresponding
integrals.  Hence $b_0^{(2)}$ is not a new independent prior: it is the joint
location--effectiveness state formed from the terminal Stage~1 beliefs.

\paragraph{Counterfactual recovery likelihood.}
Let $\{\widehat P^{(e)}\}_{e=1}^{E}$ be an ensemble of nominal world models.
At recovery round $\ell$, the method selects $q_\ell$ and observes the stacked
state response
$Y_\ell=\operatorname{vec}(s_0^{(\ell)},\ldots,s_{L_{q_\ell}}^{(\ell)})$.
Combining ensemble member $e$ with candidate dynamics $\theta$ produces
$\widehat Y_\ell^{(e)}(\theta;q_\ell)$.  The implementation uses the ensemble
mean
\begin{equation}
    \widehat Y_\ell(\theta;q_\ell)
    :=
    \frac{1}{E}\sum_{e=1}^{E}
    \widehat Y_\ell^{(e)}(\theta;q_\ell)
    \label{eq:appendix_mean_counterfactual}
\end{equation}
for recovery inference and its mean-squared trajectory discrepancy
\begin{equation}
    d_\ell(\theta)
    :=
    \operatorname{MSE}\!\left(
        Y_\ell,\widehat Y_\ell(\theta;q_\ell)
    \right),
    \label{eq:appendix_recovery_discrepancy}
\end{equation}
With a positive likelihood temperature $T_\ell$, the recovery update is
\begin{equation}
    b_{\ell+1}^{(2)}(\theta)
    \propto
    b_\ell^{(2)}(\theta)
    \exp\!\left[
        -\frac{d_\ell(\theta)}{2T_\ell}
    \right],
    \qquad T_\ell>0.
    \label{eq:appendix_normalized_recovery_update}
\end{equation}
The proportionality constant normalizes over
$\Theta_{\mathcal G}$.  In code, the smallest candidate discrepancy is
subtracted from every $d_\ell(\theta)$ before exponentiation; this common
shift is numerically stable and cancels during normalization.  The code sets
$T_\ell$ from the spread of the candidate discrepancies, subject to a
positive lower bound.
For a trajectory targeting $j(q_\ell)$, non-target fault hypotheses have the
nominal response under the single-change model.  Thus each $Y_\ell$ refines
the same joint uncertainty over location and effectiveness initialized from
Stage~1.  Ensemble-member variation is retained separately for the return
distribution below.

\appsubsection{Recovered Policy and Readiness}

\paragraph{From dynamics belief to policy.}
The terminal $b^{(2)}$ induces the random effectiveness $G_i$ defined in the
main text: $G_i=g_i$ under $H=i$ and $G_i=1$ otherwise.  Its first two moments
give the Bayes correction $c_i^\star$ in
Eq.~\eqref{eq:bayes_gain_correction}.  Applying the relevant corrections to
the nominal task policy $\pi_\kappa^0$ and clipping actions yields the
candidate recovered policy $\pi_\kappa$.  This marginalizes location and
severity uncertainty rather than committing to a hard fault estimate.

\paragraph{Predictive random return.}
For task $\kappa$, candidate policy $\pi_\kappa$, hypothesis
$\theta\sim b^{(2)}$, and ensemble index
$e\sim\operatorname{Unif}\{1,\ldots,E\}$, let
$\widehat J_\kappa^{(e)}(\pi_\kappa;\theta)$ be the return predicted by world
model $e$.  These two draws define the posterior-predictive random return
$\widehat J_\kappa$.  Its mean is
\begin{equation}
    \widehat\mu_\kappa
    =
    \frac{1}{E}
    \sum_{e=1}^{E}
    \sum_{\theta\in\Theta_{\mathcal G}}
    b^{(2)}(\theta)
    \widehat J_\kappa^{(e)}(\pi_\kappa;\theta),
    \label{eq:appendix_posterior_predictive_mean}
\end{equation}
and its variance is
\begin{equation}
    \widehat\sigma_\kappa^2
    =
    \frac{1}{E}
    \sum_{e=1}^{E}
    \sum_{\theta\in\Theta_{\mathcal G}}
    b^{(2)}(\theta)
    \left(
        \widehat J_\kappa^{(e)}(\pi_\kappa;\theta)
        -\widehat\mu_\kappa
    \right)^2.
    \label{eq:appendix_posterior_predictive_variance}
\end{equation}
By the law of total variance, this predictive variance is the sum of
world-model disagreement at fixed dynamics and uncertainty over the changed
mechanism and its effectiveness.  We use the combined quantity because both
sources can invalidate a deployment claim.

\paragraph{Readiness lower bound.}
Assume $\widehat J_\kappa$ has finite variance and the posterior-predictive
distribution is calibrated for the modeled change family.  Cantelli's
inequality states that, for every $a>0$,
\begin{equation}
    \Pr\!\left(
        \widehat J_\kappa
        <
        \widehat\mu_\kappa-a
        \,\middle|\,
        b^{(2)}
    \right)
    \leq
    \frac{\widehat\sigma_\kappa^2}
         {\widehat\sigma_\kappa^2+a^2}.
    \label{eq:appendix_cantelli_inequality}
\end{equation}
For $\widehat\sigma_\kappa>0$, setting
$a=\sqrt{(1-\alpha)/\alpha}\,\widehat\sigma_\kappa$ yields
\begin{equation}
    L_\kappa
    :=
    \widehat\mu_\kappa
    -
    \sqrt{\frac{1-\alpha}{\alpha}}\,
    \widehat\sigma_\kappa,
    \qquad
    \Pr(
        \widehat J_\kappa\geq L_\kappa
        \mid b^{(2)}
    )
    \geq1-\alpha.
    \label{eq:appendix_cantelli_lcb}
\end{equation}
If $\widehat\sigma_\kappa=0$, the predictive return is degenerate and the same
statement holds with $L_\kappa=\widehat\mu_\kappa$.
The operational interpretation is:
\begin{itemize}
    \item if $L_\kappa\geq J_{\min}$, deploy the recovered policy
    $\pi_\kappa$;
    \item if $L_\kappa<J_{\min}$, abstain and execute
    $\pi_\kappa^{\mathrm{fallback}}$; and
    \item assess calibration empirically through held-out certificate
    violations, because the bound is conditional on the modeled change family
    and predictive model.
\end{itemize}

\newpage

\appsection{Theoretical Foundations}
\label{app:theory}

\appsubsection{Proof of Proposition 1: Severity-Orthogonal Localization}

\paragraph{Proof.}
Consider the two executions in the proposition on one coupled probability
space.  Give them the same initial location belief, physical environment,
exogenous random draws, and tie-breaking draws.  The only component allowed to
differ is the rule that updates the conditional severity posteriors
$\varphi_{t,i}$.

\emph{Base case.}
By assumption, $b_0=b'_0$, and both executions have the same empty
categorical history.

\emph{Inductive step.}
Suppose that through the beginning of Stage~1 round $t$, the two executions
have the same location belief and the same categorical history.  If a
change-hazard prediction is applied before acquisition, it is the same fixed
map of the same belief, so the predicted beliefs also agree.  For every
candidate $q\in\mathcal Q$, the acquisition value in
Eq.~\eqref{eq:stage1_acquisition} depends only on that location belief, the
fixed categorical channel $p_q$, the fixed risk functional $\rho$, and the
charge $\omega_q$.  All candidate values are therefore identical across the
two executions.  The common tie-breaking rule then gives
$q_t^\star={q_t^\star}'$.

The executions now apply the same trajectory to the same coupled environment.
They consequently observe the same $\mathbf x_{q_t^\star}$.  The nominal and
calibrated-fault predictors are fixed, so the two executions also compute the
same residual, signature, matched score, and category:
\[
    \mathbf r_{q_t^\star}=\mathbf r'_{q_t^\star},
    \quad
    \mathbf h_{q_t^\star}=\mathbf h'_{q_t^\star},
    \quad
    S_{q_t^\star}=S'_{q_t^\star},
    \quad
    C_{q_t^\star}=C'_{q_t^\star}.
\]
Substituting the common category and common prior belief into
Eq.~\eqref{eq:stage1_location_update} gives
$b_{t+1}=b'_{t+1}$.  This closes the induction, so the location beliefs and
selected trajectories agree pathwise at every round up to $T$.  In
particular, their total-variation distance is zero, proving
Eq.~\eqref{eq:severity_orthogonality}.

The stopping rule is also a fixed function of the location-side state.
Because this state agrees at every round, the two executions stop at the same
time and return the same location decision.  Localization loss and detection
time are therefore equal pathwise.  Since the selected trajectory sequence is
the same, the accumulated Stage~1 charge is also equal.

The continuous coordinate also introduces no hidden interaction.  Combining
Eqs.~\eqref{eq:matched_score} and \eqref{eq:matched_coordinate} gives
\begin{equation}
    z_q
    =
    \frac{
        S_q/\lVert\mathbf h_q\rVert_2-m_{0,j(q)}
    }{
        m_{1,j(q)}-m_{0,j(q)}
    }.
    \label{eq:appendix_coordinate_reuse}
\end{equation}
Thus $z_q$ reuses the observed response and predictions already needed for
$S_q$; it requires neither another physical trajectory nor another model
rollout.  The two severity posteriors can nevertheless differ after this
shared coordinate is processed.  This does not affect the conclusion because
neither $\varphi_{t,i}$ nor any function of it is an input to the Stage~1
location update, acquisition rule, or stopping rule.
\hfill$\square$

\paragraph{Discussion.}
The proposition establishes an architectural firewall, not a calibration
claim.  Even a badly misspecified continuous severity likelihood cannot alter
which diagnostic trajectory is selected, when Stage~1 stops, or which
mechanism it reports, provided the categorical channel and location-side rules
remain fixed.  The result does not say that the transported severity posterior
is accurate; it says that any such inaccuracy is prevented from contaminating
localization.  This separation is what allows the same physical response to be
reused for recovery without paying for a second interaction or weakening the
diagnostic path.

\appsubsection{Proof of Proposition 2: Quantified Control Value}

\paragraph{Proof.}
Fix mechanism $i$ and an information set $\mathcal F$.  For any
$\mathcal F$-measurable scalar correction $c$, expand the conditional
mismatch risk as
\[
    \mathbb E[(G_i c-1)^2\mid\mathcal F]
    =
    \mathbb E[G_i^2\mid\mathcal F]c^2
    -2\mathbb E[G_i\mid\mathcal F]c+1.
\]
Completing the square and using the definition of $c_i(\mathcal F)$ in
Eq.~\eqref{eq:information_specific_risk} gives
\begin{equation}
    \mathbb E[(G_i c-1)^2\mid\mathcal F]
    =
    \mathbb E[G_i^2\mid\mathcal F]
    (c-c_i(\mathcal F))^2
    +
    1-
    \frac{\mathbb E[G_i\mid\mathcal F]^2}
         {\mathbb E[G_i^2\mid\mathcal F]}.
    \label{eq:appendix_completed_square}
\end{equation}
Because $G_i\geq g_{\min}>0$, the quadratic coefficient is positive, so
$\mathbb E[G_i^2\mid\mathcal F]\geq g_{\min}^2$ and
$c_i(\mathcal F)$ is the unique conditional Bayes correction.  The last term
in Eq.~\eqref{eq:appendix_completed_square} is its minimum conditional risk;
it is the residual control-risk factor used in
Eq.~\eqref{eq:stage2_acquisition}.

Now use $\mathcal F_0\subseteq\mathcal F_1$ and abbreviate
$c_0=c_i(\mathcal F_0)$ and $c_1=c_i(\mathcal F_1)$.  The coarse correction
$c_0$ is $\mathcal F_1$-measurable because it is already
$\mathcal F_0$-measurable.  It therefore remains a feasible decision after the
refined information is observed.  Apply
Eq.~\eqref{eq:appendix_completed_square} conditionally on $\mathcal F_1$,
first with $c=c_0$ and then with the minimizing choice $c=c_1$.  The common
minimum-risk term cancels, giving
\begin{equation}
    \mathbb E\!\left[
        (G_i c_0-1)^2-(G_i c_1-1)^2
        \,\middle|\,\mathcal F_1
    \right]
    =
    \mathbb E[G_i^2\mid\mathcal F_1](c_0-c_1)^2.
    \label{eq:appendix_conditional_risk_gap}
\end{equation}
Taking expectations and using the tower property gives the exact identity
\begin{equation}
    R_i(\mathcal F_0)-R_i(\mathcal F_1)
    =
    \mathbb E\!\left[
        \mathbb E[G_i^2\mid\mathcal F_1](c_0-c_1)^2
    \right].
    \label{eq:control_value_identity}
\end{equation}
The assumption $G_i\in[g_{\min},1]$ implies
$g_{\min}^2\leq\mathbb E[G_i^2\mid\mathcal F_1]\leq1$ almost surely.
Multiplying these pointwise bounds by the nonnegative random variable
$(c_0-c_1)^2$ and taking expectations yields
\begin{equation}
    g_{\min}^2
    \mathbb E[(c_0-c_1)^2]
    \leq
    R_i(\mathcal F_0)-R_i(\mathcal F_1)
    \leq
    \mathbb E[(c_0-c_1)^2].
    \label{eq:appendix_moment_bounds}
\end{equation}
It remains to connect the left side of
Eq.~\eqref{eq:appendix_moment_bounds} to the variance term stated in the
proposition.  Conditional on $\mathcal F_0$, add and subtract
$\mathbb E[c_1\mid\mathcal F_0]$.  The cross term has conditional expectation
zero, while $c_0$ is $\mathcal F_0$-measurable.  Hence the conditional
bias--variance decomposition yields
\begin{equation}
    \mathbb E[(c_0-c_1)^2]
    =
    \mathbb E[\operatorname{Var}(c_1\mid\mathcal F_0)]
    +
    \mathbb E[(c_0-\mathbb E[c_1\mid\mathcal F_0])^2].
    \label{eq:appendix_correction_decomposition}
\end{equation}
The second term on the right is nonnegative.  Therefore
\[
    \mathbb E[(c_0-c_1)^2]
    \geq
    \mathbb E[\operatorname{Var}(c_1\mid\mathcal F_0)].
\]
Combining this inequality with
Eqs.~\eqref{eq:appendix_moment_bounds} and
\eqref{eq:appendix_correction_decomposition} proves both bounds in
Eq.~\eqref{eq:control_value_bounds}.

Finally, the coefficient
$\mathbb E[G_i^2\mid\mathcal F_1]$ in
Eq.~\eqref{eq:appendix_conditional_risk_gap} is strictly positive almost
surely.  The risk gap is consequently zero if and only if $c_0=c_1$ almost
surely.  Equivalently, the improvement is strict exactly when the additional
gated coordinate changes the posterior-optimal correction on an event of
positive probability.
\hfill$\square$

\paragraph{Discussion.}
The result measures information by its effect on the control decision, rather
than by entropy reduction or gain-estimation error alone.  A transported
coordinate has no control value when it leaves the Bayes correction unchanged,
even if it changes other features of the posterior.  When it does change the
correction, Eq.~\eqref{eq:control_value_identity} gives the exact improvement,
and the $g_{\min}^2$ factor quantifies how residual actuator authority converts
decision refinement into mismatch-risk reduction.  The guarantee is
conditional on the gain model used to form the posterior; unlike
Proposition~\ref{prop:severity_orthogonality}, it is not robust to arbitrary
severity-model misspecification.

\appsubsection{Proof of Proposition 3: Fixed-Budget Task-Readiness Dominance}

\paragraph{Proof.}
We prove the two inequalities in Eq.~\eqref{eq:task_weighted_value}
separately.

\emph{Task-weighted control risk.}
For each mechanism, Eq.~\eqref{eq:control_value_identity} is nonnegative.
Multiplying that identity by its fixed task-importance weight $w_i\geq0$ and
summing over mechanisms gives
\begin{align}
    R_{\mathrm{ctrl}}(\mathcal F_0)
    -
    R_{\mathrm{ctrl}}(\mathcal F_1)
    &=
    \sum_{i=1}^{m}w_i
    \mathbb{E}\!\left[
        \mathbb E[G_i^2\mid\mathcal F_1]
        (c_i(\mathcal F_0)-c_i(\mathcal F_1))^2
    \right]
    \nonumber\\
    &\geq
    g_{\min}^2
    \sum_{i=1}^{m}w_i
    \mathbb{E}\!\left[
        \operatorname{Var}
        (c_i(\mathcal F_1)\mid\mathcal F_0)
    \right].
    \label{eq:appendix_task_weighted_gain}
\end{align}
Every term in the first line is nonnegative, proving
$R_{\mathrm{ctrl}}(\mathcal F_1)
\leq R_{\mathrm{ctrl}}(\mathcal F_0)$.  The second line also gives a
quantitative sufficient condition for strict improvement: some mechanism with
$w_i>0$ must have
$\operatorname{Var}(c_i(\mathcal F_1)\mid\mathcal F_0)>0$ on a set of
positive probability.

\emph{Fixed-budget task-readiness value.}
Let $\mathfrak S_{B_2}(\mathcal F)$ denote the feasible non-anticipating
Stage~2 strategies with initial information $\mathcal F$, interaction budget
$B_2$, and the action set fixed in the proposition.  A strategy specifies,
as functions of its available history, which recovery trajectory to collect,
when to stop, which task policies and certificates to construct, and when to
abstain.

Take an arbitrary coarse-information strategy
$\sigma_0\in\mathfrak S_{B_2}(\mathcal F_0)$.  Since
$\mathcal F_0\subseteq\mathcal F_1$, construct a refined-information strategy
$\widetilde\sigma_0\in\mathfrak S_{B_2}(\mathcal F_1)$ that discards the
additional coordinate and applies exactly the decision rules of $\sigma_0$.
This construction is feasible: ignoring information consumes no interaction,
does not enlarge the action set, and does not change the budget.

Couple the executions of $\sigma_0$ and $\widetilde\sigma_0$ using the same
environment and model randomness.  Initially they use the same coarse state.
Whenever their coarse histories agree, their common decision rule selects the
same next recovery trajectory.  The coupling then supplies the same recovery
observation, so their coarse histories continue to agree.  Induction over at
most $B_2$ recovery rounds shows that they collect the same trajectories,
stop at the same round, and output the same recovered policies, certificates,
and fallback decisions.  They therefore incur exactly the same
task-readiness loss on every coupled sample path.

Thus, for every loss attainable from $\mathcal F_0$, the same loss is
attainable from $\mathcal F_1$.  Minimizing over the two strategy classes gives
\[
    V_{B_2}(\mathcal F_1)
    =
    \min_{\sigma\in\mathfrak S_{B_2}(\mathcal F_1)}
    \mathbb E[\mathcal L(\sigma)]
    \leq
    \min_{\sigma\in\mathfrak S_{B_2}(\mathcal F_0)}
    \mathbb E[\mathcal L(\sigma)]
    =
    V_{B_2}(\mathcal F_0),
\]
where $\mathcal L$ is the task-readiness loss defining $V_{B_2}$.
Proposition~\ref{prop:severity_orthogonality} additionally ensures that
admitting the gated coordinate does not change the Stage~1 path that precedes
these strategy classes.

A useful sufficient condition for strictness is the following.  Suppose that,
on an event of positive probability, the refined-information Bayes strategy
has a unique optimal recovery or deployment decision, every
$\mathcal F_0$-measurable decision differs from it on that event, and the
resulting conditional loss gap is positive.  A coarse strategy must use one
decision across refined subevents that it cannot distinguish, so it incurs
that positive gap with positive probability.  Then
$V_{B_2}(\mathcal F_1)<V_{B_2}(\mathcal F_0)$.
\hfill$\square$

\paragraph{Discussion.}
This proposition lifts the one-step control-risk result to the complete
fixed-budget decision problem.  Its central message is a value-of-information
statement: an exact Bayes agent can always ignore transported evidence, so
having that evidence cannot make the best achievable readiness decision
worse.  The comparison is fair only because both agents have the same physical
interaction budget and action set, and the transported coordinate is obtained
from an already-paid-for Stage~1 response.  The inequality concerns optimal
non-anticipating strategies under a calibrated joint model.  It does not claim
that every greedy acquisition heuristic improves on every sample path, nor
that extra information remains beneficial under an arbitrarily misspecified
posterior.

\paragraph{Overall scope.}
Proposition~\ref{prop:severity_orthogonality} is exact, whereas
Propositions~\ref{prop:control_value}--\ref{prop:end_to_end_value} are
model-conditional Bayes results; world-model, gain-grid, and greedy-acquisition
approximation effects are evaluated empirically.

\newpage
\appsection{EG-MPT Algorithmic Specification}
\label{app:egmpt_algorithm}

This appendix summarizes EG-MPT as an executable two-stage procedure.  It
gives only the notation needed to follow the information flow, followed by
pseudocode and a step-wise explanation.  Calibration-only and proof-local
quantities are deliberately defined where they are used instead of being
promoted to global notation.

\appsubsection{Core Notation and Interface}

Offline inputs are the diagnostic library, held-out categorical channels and
response statistics, counterfactual predictors, gain grid, task weights, and
nominal policies.  Online inputs are passive observations, the two trajectory
budgets, and the readiness thresholds.  The method outputs, for every candidate
task, a recovered policy, a continuous readiness lower bound, and the policy
actually deployed after possible abstention.

\begin{table}[h]
\centering
\caption{Core notation.  Symbols local to one derivation or
implementation detail are defined in place and omitted here.}
\label{tab:egmpt_notation}
\small
\setlength{\tabcolsep}{4pt}
\renewcommand{\arraystretch}{1.12}
\begin{tabular}{|p{0.12\textwidth}|p{0.34\textwidth}|
                p{0.12\textwidth}|p{0.34\textwidth}|}
\hline
Symbol & Meaning & Symbol & Meaning \\
\hline
$H$ & Changed mechanism; $H=0$ is nominal.
& $g_i,G_i$ & Realized and posterior-random effectiveness of mechanism $i$.\\
$\mathcal K,\kappa,\nu$ & Candidate tasks, revealed task, and reveal
distribution.
& $\mathcal Q,q,j(q)$ & Diagnostic library, one trajectory, and its target.\\
$q_t,q_\ell$ & Trajectories selected at Stage~1 round $t$ and Stage~2 round
$\ell$.
& $\mathbf x_{q_t},Y_\ell$ & Stage~1 diagnostic response and Stage~2 recovery
response.\\
$b_t(h)$ & Stage~1 posterior over $H$.
& $\varphi_{t,i}(g)$ & Effectiveness posterior conditional on $H=i$.\\
$\mathbf r_q,\mathbf h_q$ & Response residual and calibrated fault signature.
& $S_q,C_q,z_q$ & Matched score, its category, and normalized amplitude.\\
$b_\ell^{(2)}$ & Joint location--effectiveness recovery posterior.
& $B_1,B_2,\omega_q$ & Stage budgets and trajectory charge.\\
$w_i,c_i^\star$ & Future-task importance and Bayes action correction.
& $\pi_\kappa,L_\kappa$ & Recovered policy and readiness lower bound.\\
$\alpha,J_{\min}$ & Certificate error level and required return.
& $\pi_\kappa^{\mathrm{dep}}$ & Recovered-or-fallback deployed policy.\\
\hline
\end{tabular}
\end{table}

\paragraph{Symbol flow.}
The notation in Table~\ref{tab:egmpt_notation} follows three consecutive
transformations.  The Stage~1 diagnostic update is
\[
    \boxed{
    q_t
    \ \longrightarrow\
    \mathbf x_{q_t}
    \ \longrightarrow\
    (S_{q_t},C_{q_t},z_{q_t})
    \ \longrightarrow\
    \bigl(b_{t+1},\{\varphi_{t+1,i}\}_{i=1}^{m}\bigr)
    }.
\]
Here $C_{q_t}$ updates $b_{t+1}$, while $z_{q_t}$ updates only the targeted
conditional posterior when the gate opens.  The stage boundary and recovery
recursion are
\[
    \boxed{
    \bigl(b^{(1)},\{\varphi_i^{(1)}\}_{i=1}^{m}\bigr)
    \ \longrightarrow\
    b_0^{(2)}
    \ \xrightarrow{\ (q_\ell,Y_\ell)\ }\
    b_{\ell+1}^{(2)}
    \ \longrightarrow\
    b^{(2)}
    }.
\]
Finally, the terminal joint belief is converted into task-wise actions and
decisions:
\[
    \boxed{
    b^{(2)}
    \ \longrightarrow\
    \{c_i^\star\}_{i=1}^{m}
    \ \longrightarrow\
    \pi_\kappa
    \ \longrightarrow\
    (\widehat\mu_\kappa,\widehat\sigma_\kappa,L_\kappa)
    \ \longrightarrow\
    \pi_\kappa^{\mathrm{dep}}
    }.
\]
These boxes distinguish three roles that are easy to conflate: localization
state $b_t$, conditional effectiveness state $\varphi_{t,i}$, and joint
recovery state $b_\ell^{(2)}$.  Only the last state is used for recovery
acquisition, policy correction, and certification.
\FloatBarrier

\appsubsection{End-to-End Pseudocode}

Algorithms~\ref{alg:egmpt_stage1}--\ref{alg:egmpt_stage2} make explicit the
single posterior state passed between the two stages.  All calibration
quantities are fixed before evaluation; only beliefs and observed responses
are updated online.

\begin{algorithm}[t]
\caption{Stage~1: localization and evidence-gated transport}
\label{alg:egmpt_stage1}
\small
\begin{algorithmic}[1]
\REQUIRE $\mathcal Q$, calibrated predictors, $p_q$, $m_{0,j}$, $m_{1,j}$,
$\sigma_j$, gain grid $\mathcal G$, risk $\rho$, charges $\omega_q$, budget
$B_1$
\ENSURE transported recovery posterior $b_0^{(2)}$ and the Stage~1 stop
decision
\STATE Initialize $b_0$ and each $\varphi_{0,i}$
\FOR{$t=0,\ldots,B_1-1$}
    \STATE Apply the change-hazard prediction to $b_t$
    \STATE Select
    \[
        q_t\in\arg\max_{q\in\mathcal Q}
        \frac{\rho(b_t)-
        \mathbb E_{C_q\sim p(\cdot\mid b_t,q)}
        [\rho(b_{t+1}^{\,q,C_q})]}{\omega_q}
    \]
    \STATE Execute $q_t$; compute $\mathbf r_{q_t}$, $S_{q_t}$, $C_{q_t}$,
    and $z_{q_t}$
    \STATE Update $b_{t+1}$ by Eq.~\eqref{eq:stage1_location_update}
    \IF{$p_{q_t}(C_{q_t}\mid j(q_t))>p_{q_t}(C_{q_t}\mid0)$}
        \STATE Update $\varphi_{t+1,j(q_t)}$ by
        Eq.~\eqref{eq:gated_gain_update}
    \ELSE
        \STATE Leave the targeted conditional posterior unchanged
    \ENDIF
    \STATE Leave all non-targeted conditionals unchanged
    \IF{the fixed stopping rule fires on $b_{t+1}$}
        \STATE \textbf{break}
    \ENDIF
\ENDFOR
\STATE Let $b^{(1)}$ and $\varphi_i^{(1)}$ be the terminal posteriors
\STATE Form $b_0^{(2)}$ using
Eqs.~\eqref{eq:recovery_initialization} and
\eqref{eq:transported_initialization}
\STATE \textbf{return} $b_0^{(2)}$ and whether stopping fired
\end{algorithmic}
\end{algorithm}

\begin{algorithm}[t]
\caption{Stage~2: recovery, policy construction, and readiness}
\label{alg:egmpt_stage2}
\small
\begin{algorithmic}[1]
\REQUIRE $b_0^{(2)}$, Stage~1 stop decision, $\mathcal Q$, counterfactual
model ensemble, $\mathcal G$, weights $w_i$, charges $\omega_q$, $B_2$,
$\alpha$, $J_{\min}$
\ENSURE $\{(\pi_\kappa,L_\kappa,\pi_\kappa^{\mathrm{dep}})
\}_{\kappa\in\mathcal K}$
\IF{Stage~1 stopping fired}
    \FOR{$\ell=0,\ldots,B_2-1$}
        \STATE Select
        \[
            q_\ell\in\arg\max_{q\in\mathcal Q}
            \frac{w_{j(q)}
            \left(1-
            \mathbb E[G_{j(q)}\mid b_\ell^{(2)}]^2/
            \mathbb E[G_{j(q)}^2\mid b_\ell^{(2)}]\right)}
            {\omega_q}
        \]
        \STATE Execute $q_\ell$ and observe $Y_\ell$
        \STATE Update $b_{\ell+1}^{(2)}$ using
        Eq.~\eqref{eq:appendix_normalized_recovery_update}
    \ENDFOR
    \STATE Set $b^{(2)}\gets b_{B_2}^{(2)}$
\ELSE
    \STATE Set $b^{(2)}\gets b_0^{(2)}$; collect no Stage~2 trajectory
\ENDIF
\FOR{each $\kappa\in\mathcal K$}
    \STATE Compute the relevant corrections $c_i^\star$ using
    Eq.~\eqref{eq:bayes_gain_correction}
    \STATE Construct $\pi_\kappa$ from $\pi_\kappa^0$ and clip its actions
    \STATE Predict return under all posterior hypotheses and ensemble members
    \STATE Compute $\widehat\mu_\kappa$, $\widehat\sigma_\kappa$, and
    $L_\kappa$ using
    Eqs.~\eqref{eq:appendix_posterior_predictive_mean}--%
    \eqref{eq:appendix_posterior_predictive_variance} and
    Eq.~\eqref{eq:cantelli_certificate}
    \STATE Set $\pi_\kappa^{\mathrm{dep}}$ by
    Eq.~\eqref{eq:deployment_rule}
\ENDFOR
\STATE \textbf{return}
$\{(\pi_\kappa,L_\kappa,\pi_\kappa^{\mathrm{dep}})
\}_{\kappa\in\mathcal K}$
\end{algorithmic}
\end{algorithm}

\paragraph{Stage boundary.}
The location marginal and conditional effectiveness distributions are two
views of one inferential state.  Their product, with nominal point masses on
unaffected mechanisms, forms $b_0^{(2)}$; no hard diagnosis is inserted and
Stage~2 does not restart from a broad prior.  Algorithm~\ref{alg:egmpt_stage2}
can therefore use the transported state immediately, or retain it unchanged
on the no-alarm branch.

\paragraph{Termination and decision semantics.}
The procedure returns a triple for every possible future task, not merely a
binary recovery decision.  The candidate policy $\pi_\kappa$ records the
action correction implied by the posterior, while $L_\kappa$ reports the
continuous evidence-supported lower return.  Only
$\pi_\kappa^{\mathrm{dep}}$ applies the threshold $J_{\min}$: it deploys the
candidate when the readiness claim is sufficient and otherwise selects the
fallback.  This separation makes policy quality and confidence independently
auditable.

\appsubsection{Step-by-Step Interpretation}

\begin{enumerate}
    \item \textbf{Calibrate response geometry.}
    For each $q$, nominal and calibrated-fault rollouts determine the signature
    $\mathbf h_q$, categorical probabilities $p_q$, robust centers, and
    dispersion.  These quantities are
    estimated on calibration trajectories disjoint from evaluation data.  The
    separation is necessary because reusing evaluation outcomes to choose bins
    or likelihood scales would make both detection and certificate metrics
    optimistically biased.

    \item \textbf{Probe for localization value.}
    Stage~1 selects the trajectory with the greatest expected reduction in
    task-weighted location risk per unit charge.  From the same physical
    response, the inner product
    $\langle\mathbf r_q,\mathbf h_q\rangle$ yields both $S_q$ and $z_q$.
    Under the local model
    $\mathbf r_q=\beta_q\mathbf h_q+\boldsymbol\epsilon_q$ with isotropic
    residual noise, the likelihood ratio for positive amplitude is monotone in
    this inner product and
    $\langle\mathbf r_q,\mathbf h_q\rangle/
    \lVert\mathbf h_q\rVert_2^2$ is the least-squares and maximum-likelihood
    estimate of $\beta_q$.
    Normalization by $\lVert\mathbf h_q\rVert_2$ gives $S_q$ a stable noise
    scale for detection, while normalization by
    $\lVert\mathbf h_q\rVert_2^2$ makes $z_q$ an amplitude estimate for gain
    recovery.  No second physical trajectory or duplicate model rollout is
    introduced.

    \item \textbf{Separate localization from severity.}
    The category $C_q$ alone updates $b_t$.  If that category favors the
    targeted fault over nominal dynamics, $z_q$ also updates
    $\varphi_{t,j(q)}$; otherwise the conditional effectiveness posterior is
    unchanged.  The positive-likelihood-ratio gate maximizes, among
    deterministic gates, the difference between its opening probabilities
    under the targeted fault and nominal dynamics
    \citep{neyman1933efficient}.  Thus an inaccurate gain model cannot change the trajectory
    sequence, stopping path, or location belief, as formalized by
    Proposition~\ref{prop:severity_orthogonality}.

    \item \textbf{Transport instead of restarting.}
    At the stage boundary, $b^{(1)}$ supplies the location marginal and
    $\varphi_i^{(1)}$ supplies the conditional gain distribution.  Their product
    forms $b_0^{(2)}$.  Recovery then spends its limited interactions where
    posterior mismatch risk, future-task importance, and interaction charge
    make another trajectory most valuable.  Counterfactual responses refine
    the same joint posterior.

    \item \textbf{Recover and certify.}
    The correction $c_i^\star$ minimizes posterior expected squared actuation
    mismatch.  The world-model ensemble and $b^{(2)}$ induce a predictive return
    distribution for every candidate task.  Its Cantelli lower bound
    $L_\kappa$ remains continuous; thresholding occurs only when it is compared
    with $J_{\min}$.  If the bound is insufficient, the method abstains rather
    than converting uncertainty into an unsupported deployment claim.
\end{enumerate}

\paragraph{No-alarm branch.}
Reaching $B_1$ without an alarm does not assert that the system is certainly
nominal.  The default protocol skips costly Stage~2 recovery but retains the
posterior uncertainty in $b_0^{(2)}$.  That uncertainty propagates into
$L_\kappa$ and can still force abstention for tasks whose return is not
adequately supported.

\appsubsection{Implementation Invariants and Computational Footprint}

The pseudocode exposes four invariants that should hold in any implementation:
\begin{itemize}
    \item \textbf{No severity feedback:} continuous-effectiveness updates never
    enter the Stage~1 location update, acquisition rule, or stopping rule.
    \item \textbf{No hidden interaction charge:} $S_q$, $C_q$, and $z_q$ are
    computed from the same observed diagnostic response.
    \item \textbf{One recovery state:} Stage~2 acquisition, policy correction,
    predictive return, and readiness all use the same $b^{(2)}$.
    \item \textbf{Task-agnostic physical recovery:} all Stage~2 interactions
    occur before task identity is revealed; revealing $\kappa$ selects from the
    prepared policy/certificate family without additional recovery data.
\end{itemize}

Excluding model simulation, one Stage~1 acquisition pass uses
$O(|\mathcal Q|K|\mathcal H|)$ categorical operations, and an open gate costs
$O(|\mathcal G|)$.  A Stage~2 update evaluates one counterfactual response per
ensemble member and location--gain hypothesis; certification additionally
evaluates each candidate task.  In practice, these model rollouts dominate the
finite-support arithmetic and parallelize across hypotheses and ensemble
members.

\newpage
\appsection{Experimental Settings for Main Results}
\label{app:experimental_details}

This section records the settings for the experiments reported in the main
text.  Its organization follows the main-text experimental section exactly:
the common protocol is specified first, followed by Readiness under Scarce
Evidence, Severity and Transfer, Decision Quality, and Breadth and Risk.
Unless an experiment-specific override is stated below, all methods use the
reference protocol in Tables~\ref{tab:environment_task_settings}
and~\ref{tab:shared_experimental_settings}.  Thus each comparison changes only
the factor named in its paragraph.

\appsubsection{Common Protocol and Baseline Adaptations}

\paragraph{Dormant-actuator MuJoCo setting.}
We use the Gymnasium v5 versions of Ant, HalfCheetah, Hopper, Humanoid,
Swimmer, and Walker2d as controlled transition systems.  The downstream
objective is actuator-indexed trajectory tracking rather than the native
locomotion reward, which makes recovery quality comparable to a known nominal
reference.  Let \(d_a\) denote the action dimension and let
\(a_{\mathrm{cur}}=0\) be the actuator used by the pre-reveal behavior.  The
candidate changed actuators are \(a\in\{1,\ldots,d_a-1\}\).  Passive
trajectories excite only \(a_{\mathrm{cur}}\), so multiplying any candidate
actuator by a gain \(g<1\) leaves the passive action sequence unchanged and
realizes the dormancy condition in Eq.~\eqref{eq:dormant_equivalence}.

A trial is nominal with probability \(0.2\).  Otherwise, one candidate
actuator is sampled uniformly and its applied command is multiplied by
\(g_{\mathrm{true}}=0.35\).  The hidden change round is uniform on
\(\{10,\ldots,20\}\), and the future task is revealed at a later hidden round
uniform on \(\{35,\ldots,45\}\).  A future task requests the nominal trajectory
generated by pulsing one candidate actuator.  Table~\ref{tab:environment_task_settings}
specifies the reveal distribution and relative importance of these tasks;
candidate entries follow increasing action index.

\begin{table}[H]
\centering
\caption{Environment-specific future-task settings in the reference MuJoCo
protocol.  The task-aware acquisition weight is \(w_a=\nu_a s_a\), where
\(\nu_a\) is the reveal probability and \(s_a\) is the relative
control-importance multiplier.}
\label{tab:environment_task_settings}
\scriptsize
\setlength{\tabcolsep}{4pt}
\renewcommand{\arraystretch}{1.08}
\begin{tabular}{|l|c|c|p{0.30\textwidth}|p{0.27\textwidth}|}
\hline
Environment & \(d_a\) & Candidates & \(\nu\) & \(s\) \\
\hline
Ant & 8 & 7 &
\((.08,.10,.12,.14,.16,.18,.22)\) &
\((1,1.1,1.2,1.4,1.6,1.8,2)\) \\
HalfCheetah & 6 & 5 &
\((.10,.15,.20,.25,.30)\) &
\((1,1.2,1.4,1.7,2)\) \\
Hopper & 3 & 2 & \((.50,.50)\) & \((1,1)\) \\
Humanoid & 17 & 16 & uniform \(1/16\) & all \(1\) \\
Swimmer & 2 & 1 & \((1)\) & \((1)\) \\
Walker2d & 6 & 5 &
\((.10,.15,.20,.25,.30)\) &
\((1,1.2,1.4,1.7,2)\) \\
\hline
\end{tabular}
\end{table}

\paragraph{Shared budgets, calibration, and models.}
A \emph{physical trajectory} is one actual environment rollout from a saved
reset state; counterfactual world-model rollouts do not consume this budget.
Calibration and evaluation trajectories are disjoint.  Every method receives
the same labelled calibration set, although its diagnostic statistic induces
its own held-out categorical channel.  The reference values are collected in
Table~\ref{tab:shared_experimental_settings}.

\begin{table}[t]
\centering
\caption{Reference MuJoCo experimental and implementation settings.}
\label{tab:shared_experimental_settings}
\small
\setlength{\tabcolsep}{5pt}
\renewcommand{\arraystretch}{1.08}
\begin{tabular}{|p{0.39\textwidth}|p{0.53\textwidth}|}
\hline
Setting & Value \\
\hline
Evaluation repetitions & seeds \(0\)--\(4\); 50 trials per seed and environment \\
Trial mixture & 20\% nominal; 80\% one dormant-actuator change \\
Reference changed gain & \(g_{\mathrm{true}}=0.35\) \\
Change / task-reveal rounds & uniform on \([10,20]\) / uniform on \([35,45]\) \\
Passive trajectory & horizon 8, amplitude 0.20 on \(a_{\mathrm{cur}}\) \\
Stage~1 diagnostic trajectory & horizon 8, amplitude 0.25; one opportunity
every 5 rounds \\
Interaction charge & 0.01 per diagnostic step \\
Stage~2 physical budget & at most \(B_2=2\) recovery trajectories, after an alert \\
Future tracking task & horizon 12, amplitude 0.25 \\
Labelled channel calibration & 100 episodes; \(K=5\) score bins; additive smoothing 1 \\
Alert calibration & 2,000 nominal trials; 5\% anytime family-wise target \\
Reveal-decision calibration & 2,000 independent trials under the shared
mitigation loss \\
Gain support & 9-point grid on \([0.15,0.95]\), augmented with \(0.35\) \\
Recovery likelihood temperature & relative temperature 0.05 \\
Readiness rule & \(\alpha=0.10\), \(J_{\min}=0.85\) \\
Dynamics data & 500,000 colored-noise transitions per environment \\
World-model ensemble & 12 members; three 512-unit hidden layers \\
World-model optimization & 300 epochs, batch size 1,024, early-stopping
patience 30 \\
\hline
\end{tabular}
\end{table}

\paragraph{Metrics and aggregation.}
For future task \(\kappa\), let \(\mathrm{NTE}_\kappa\) be trajectory-tracking
error normalized by the nominal reference and clipped at 10.  The return and
selective return are
\begin{equation}
    R_\kappa := \exp(-\mathrm{NTE}_\kappa),
    \qquad
    R_{\mathrm{sel}}
    := \sum_{\kappa\in\mathcal K}
       \nu_\kappa R_\kappa^{\mathrm{dep}},
    \label{eq:appendix_selective_return}
\end{equation}
where \(R_\kappa^{\mathrm{dep}}\) evaluates the corrected policy when its
readiness lower bound is at least \(J_{\min}\), and otherwise evaluates the
nominal fallback.  Detection metrics are evaluated only on changed trials,
whereas selective return uses the complete stratified mixture.  Each metric is
first averaged over trials within a seed; reported uncertainty is the sample
standard deviation of the five seed-level means.  Regret, gain error,
certificate violations, delay, and interaction cost are lower-is-better;
return and detection rate are higher-is-better.

\paragraph{Baseline adaptations.}
The source methods assume different interfaces.  To isolate acquisition
quality, we retain each defining selection principle but map every method to
the same finite diagnostic-trajectory library:
\begin{itemize}
    \item \textbf{Bandit-QCD} maintains a nonnegative cumulative
    log-likelihood statistic for every candidate actuator and adds an
    exploration bonus when selecting the next sensing arm
    \citep{gopalan2021bandit}.
    \item \textbf{SEPT} ranks trajectories once by symmetric categorical
    information per interaction charge and repeats the resulting open-loop
    schedule \citep{yang2020sept}.
    \item \textbf{ASID-FIM} uses the finite secant between nominal and
    calibrated-fault categorical channels as a local likelihood derivative,
    then selects the largest A-optimal reduction in posterior covariance
    \citep{memmel2024asid}.
    \item \textbf{OPAX} selects task-agnostic mutual information per
    interaction charge \citep{sukhija2023opax}.
    \item \textbf{Task-OED} multiplies identification information by posterior
    uncertainty and future-task control weight
    \citep{wagenmaker2023optimal}.
    \item \textbf{Bayes-Risk} selects the largest expected reduction in
    task-weighted posterior risk per interaction charge
    \citep{blackwell1953equivalent}; \textbf{Random} samples uniformly.
\end{itemize}
All methods share the alert and mitigation thresholds, physical budgets,
Stage~2 likelihood, Bayes correction, readiness certificate, and fallback.
Baselines begin Stage~2 from the broad gain prior; EG-MPT begins from its gated
transported Stage~1 coordinate.  Transport therefore receives neither an
extra trajectory nor additional labelled calibration.

\appsubsection{Readiness under Scarce Evidence}

\paragraph{Calibration sensitivity (Figure~\ref{fig:exp06_sensitivity}).}
The sensitivity study uses Ant, HalfCheetah, Hopper, and Walker2d.  In the
categorical-resolution sweep, \(K\in\{2,3,5,9\}\) with 100 labelled
calibration episodes.  In the calibration-data sweep,
\(N\in\{25,50,100,200\}\) with \(K=5\).  The \(K=5,N=100\) configuration is
the shared reference point; all other settings in
Table~\ref{tab:shared_experimental_settings} remain fixed.

\paragraph{End-to-end readiness (Table~\ref{tab:exp01_readiness}).}
The primary readiness comparison uses all six MuJoCo environments and the
reference protocol without overrides.  It reports detection, diagnostic
interaction, gain estimation, recovery interaction, readiness, certificate,
and deployed-control quantities from the same trials, so intermediate
diagnostic performance and final deployment performance are not evaluated on
different trial populations.

\paragraph{Recovery budgets and deadlines
(Table~\ref{tab:exp04_05_budget_deadline}).}
The budget sweep uses Ant, HalfCheetah, Hopper, and Walker2d and sets
\(B_2\in\{0,1,2,4\}\); the main table displays \(B_2\in\{0,1,2\}\).
The deadline sweep uses Ant, HalfCheetah, Walker2d, and Humanoid.  It fixes the
change at round 10, offers one diagnostic opportunity every five rounds, and
sets the reveal round to \(10+5P\), where
\(P\in\{1,2,4,6\}\) is the number of post-change opportunities.  The main
table displays \(P\in\{2,4,6\}\).  The two sweeps vary physical recovery
evidence and pre-reveal diagnostic evidence separately.

\appsubsection{Severity and Transfer}

\paragraph{Severe faults (Table~\ref{tab:exp02_severe_faults}).}
Ant, HalfCheetah, Hopper, and Walker2d are evaluated at
\(g_{\mathrm{true}}\in\{0.20,0.35,0.50,0.65,0.80\}\).  In this
\emph{in-distribution} sweep, the labelled fault channel is recalibrated at
each tested gain, i.e., \(g_{\mathrm{cal}}=g_{\mathrm{true}}\).  The main
table shows the severe \(g=0.20\) and reference \(g=0.35\) cases.  This
distinction is important: unlike the unseen-severity experiment below, this
study changes fault magnitude without introducing calibration shift.

\paragraph{Cross-physics transfer (Table~\ref{tab:exp10_cross_physics}).}
The main result aggregates BipedalWalker-v3 and CarRacing-v3 under the same
dormant-effector construction.  BipedalWalker uses four continuous joint
commands, with actuator 0 current and actuators \(1,2,3\) as candidates; its
eight-step diagnostic uses zero-integral pulses of amplitude 0.55.  CarRacing
uses the steering, throttle, and brake commands; throttle (index 1) defines the
current behavior and steering/brake are candidates.  After 24 throttle warm-up
steps, the eight-step diagnostics use steering amplitude 0.65 or brake
amplitude 0.80.  Both backends use five categorical bins, calibration gain
0.35, a 17-point gain grid on \([0.15,0.95]\), coordinate-noise scale 0.04,
\(B_2=1\), and 50 trials for each of seeds \(0\)--\(4\).  True gains are
\(0.30,0.50,0.70\), and the main table reports all three settings.

\paragraph{No recovery interaction
(Table~\ref{tab:exp09v2_severity_transport}).}
The native Gymnasium Pendulum-v1 and Acrobot-v1 tasks isolate the Stage~1
severity belief by fixing \(B_2=0\).  Each backend uses its native
\texttt{reset}/\texttt{step} dynamics and native control score.  True gains
are \(0.35,0.50,0.65\), with the main table reporting \(0.35\) and \(0.50\).
There are 30 trials per method, environment, and seed for seeds \(0\)--\(4\).
Because no recovery trajectory is permitted, any gain-error or CRPS difference
must be present before additional Stage~2 interaction.

\appsubsection{Decision Quality}

\paragraph{Evidence gate (Table~\ref{tab:expN7_transport_ablation}).}
The ablation uses the same unseen-severity Box2D suite described below:
BipedalWalker-v3 and BipedalWalkerHardcore-v3, calibration gain 0.35, unseen
true gains \(0.20,0.50,0.65\), and \(B_2\in\{0,1\}\).  The
\textsc{Ours--no-gate} variant differs from EG-MPT only by removing the
positive-evidence condition before transporting the matched coordinate.
Trajectory libraries, calibration, budgets, Stage~2 inference, certification,
and aggregation are unchanged.  The comparison therefore attributes changes
in deployment, severity error, and violations to the gate rather than to a
different recovery policy.

\paragraph{Mechanism diagnostics (Table~\ref{tab:exp04_diagnostics}).}
These diagnostics are computed from the MuJoCo recovery-budget sweep at the
shared \(B_2=2\) setting on Ant, HalfCheetah, Hopper, and Walker2d.  The
reported gain error, detection rate, violation rate, and delay are taken from
the same runs as the end-to-end budget result.  Gain error is the absolute
error of the posterior mean for the changed actuator; a certificate violation
occurs when a deployed task realizes return below its certified lower bound;
detection delay counts rounds from the hidden change to the first alert.
These definitions expose the localization, severity-estimation, and
certification decisions that are compressed into selective return.

\appsubsection{Breadth and Risk}

\paragraph{Compact plants (Table~\ref{tab:expM1_estimation}).}
The compact-plant suite contains eleven families: over-actuated attitude
control, multi-zone building thermal control, a pump--tank network, parallel
battery strings, a satellite reaction-wheel array, aircraft short-period
control with redundant surfaces, a Wood--Berry distillation column, a DC
servo, a quadruple-tank process, quadrotor altitude control, and wind-turbine
pitch control.  The main table displays six representative families, while
all eleven are run under the same pre-specified matrix.  Each family reserves
one current effector and treats every other effector as dormant; numerical
invariance is checked before evaluation.

\begin{table}[t]
\centering
\caption{Compact-plant evaluation settings.}
\label{tab:expM1_settings}
\small
\setlength{\tabcolsep}{5pt}
\renewcommand{\arraystretch}{1.08}
\begin{tabular}{|p{0.30\textwidth}|p{0.61\textwidth}|}
\hline
Axis & Values \\
\hline
Plant families & Attitude, building thermal, pump--tank, battery,
reaction wheel, aircraft pitch, Wood--Berry, DC servo, quadruple tank,
quadrotor, wind pitch \\
Methods & Random, Bayes-Risk, Bandit-QCD, SEPT, ASID-FIM, OPAX, Task-OED,
EG-MPT \\
True gain / recovery budget & \(g\in\{0.20,0.50\}\); \(B_2=1\) \\
Evaluation repetitions & seeds \(0\)--\(4\); 40 trials per seed and family \\
Channel calibration & \(g_{\mathrm{cal}}=0.35\); 60 labelled episodes;
5 bins; coordinate-noise scale 0.06 \\
Gain support & 17-point grid on \([0.15,0.95]\) \\
\hline
\end{tabular}
\end{table}

The two main-table quantities are \(\lvert c_j g_j-1\rvert\), the
compensation error delivered to the plant, and the largest posterior-mean
effectiveness error over dormant mechanisms.  No additional result table is
included here; Table~\ref{tab:expM1_settings} specifies the complete
eleven-family evaluation matrix referenced by the main text.

\paragraph{Unseen severity (Table~\ref{tab:expN7_unseen_severity}).}
BipedalWalker-v3 and BipedalWalkerHardcore-v3 are calibrated only at
\(g_{\mathrm{cal}}=0.35\), then evaluated at
\(g_{\mathrm{true}}\in\{0.20,0.50,0.65\}\), none of which occurs in the
categorical calibration channel.  Both \(B_2=0\) and \(B_2=1\) are run, with
50 trials per method, environment, and seed over seeds \(0\)--\(4\).  The
gain support remains the 17-point grid on \([0.15,0.95]\).  This fixed
calibration anchor is the defining difference from the in-distribution
severity sweep in Table~\ref{tab:exp02_severe_faults}.

\paragraph{Future-task risk (Table~\ref{tab:exp07_risk_profiles}).}
Ant, HalfCheetah, and Walker2d use three reveal profiles while all physical and
statistical budgets remain fixed.  The uniform profile assigns equal
probability and unit importance to every candidate task.  The head-heavy
profile assigns probabilities proportional to
\((d_a-1,d_a-2,\ldots,1)\) and unit importance.  The rare-critical profile
uses the same decreasing probabilities but assigns the least likely, last
candidate importance \(2d_a\), leaving all others at one.  In every case the
acquisition weight is \(w_a=\nu_a s_a\).  Hence this experiment changes future
consequence, not diagnostic access or fault severity.

\FloatBarrier
\newpage
\appsection{Additional Experimental Results}
\label{app:additional_experimental_results}

This section reports completed experiments that supplement, rather than define,
the main comparisons.  Settings are stated with each experiment because these
studies deliberately change different parts of the reference protocol in
Appendix~\ref{app:experimental_details}.  Tables are reproduced directly here;
the appendix does not depend on external \texttt{.tex} fragments.

\appsubsection{Additional MuJoCo Controls}

\paragraph{Transfer from one calibration fault to unseen MuJoCo severities.}
We fix the categorical calibration gain at \(g_{\mathrm{cal}}=0.35\) and
evaluate Ant, HalfCheetah, Walker2d, and Humanoid at
\(g_{\mathrm{true}}\in\{0.20,0.50,0.65,0.80\}\).  All eight methods use
seeds \(0\)--\(4\), 50 trials per environment and seed, and the reference
\(B_2=2\) recovery budget.  This differs from the in-distribution sweep in
Table~\ref{tab:exp02_severe_faults}, where the channel is recalibrated at every
tested gain.

The transported coordinate generally improves the correction on changed tasks,
especially at severe and moderate loss, but that advantage does not carry to
mixture-level regret in this early MuJoCo transfer protocol
(Table~\ref{tab:additional_mujoco_controls}).  The discrepancy is informative:
a sharper conditional severity estimate cannot compensate for every
localization, alert, and fallback error accumulated over the full nominal/fault
mixture.  This control therefore motivates evaluating both the changed-task
correction and the complete deployment decision.

\paragraph{Matched-response strength.}
On Ant, HalfCheetah, and Walker2d, we sweep diagnostic pulse amplitude over
\(\{0.10,0.15,0.25,0.35\}\), retaining \(g_{\mathrm{true}}=
g_{\mathrm{cal}}=0.35\), five seeds, 50 trials per environment and seed, and
all remaining reference settings.  Stronger excitation steadily reduces gain
error, while changed-task return largely saturates by amplitude \(0.25\).
Mixture regret again does not improve monotonically, because stronger
responses help severity estimation without changing the alert and fallback
logic.  The reference amplitude \(0.25\) is consequently a compromise: it
captures most of the correction benefit without relying on the most aggressive
diagnostic pulse.

\begin{table}[H]
\centering
\caption{Additional MuJoCo controls.  Entries are macro means over environments
and five seeds.  ``Best base'' is the strongest of the seven baselines,
selected separately for each metric.}
\label{tab:additional_mujoco_controls}
\scriptsize
\setlength{\tabcolsep}{3.8pt}
\renewcommand{\arraystretch}{1.05}
\begin{tabular}{|l|c|c|c|c|c|c|}
\hline
& \multicolumn{2}{c|}{Changed-task return \(\uparrow\)}
& \multicolumn{2}{c|}{Gain MAE \(\downarrow\)}
& \multicolumn{2}{c|}{Mixture regret \(\downarrow\)} \\
\hline
Setting & EG-MPT & Best base & EG-MPT & Best base & EG-MPT & Best base \\
\hline
\multicolumn{7}{|l|}{\textit{Fixed calibration, unseen true gain}} \\
\hline
\(g=0.20\) & \underline{$.771$} & $.746$ & \underline{$.353$} & $.410$ & $.808$ & \underline{$.764$} \\
\(g=0.50\) & \underline{$.935$} & $.906$ & \underline{$.238$} & $.263$ & $.769$ & \underline{$.722$} \\
\(g=0.65\) & \underline{$.958$} & $.937$ & $.169$ & \underline{$.162$} & $.760$ & \underline{$.709$} \\
\(g=0.80\) & \underline{$.971$} & $.966$ & \underline{$.116$} & $.133$ & $.753$ & \underline{$.688$} \\
\hline
\multicolumn{7}{|l|}{\textit{Matched-response pulse amplitude}} \\
\hline
\(A=0.10\) & \underline{$.889$} & $.803$ & \underline{$.258$} & $.348$ & $.787$ & \underline{$.751$} \\
\(A=0.15\) & \underline{$.906$} & $.812$ & \underline{$.233$} & $.329$ & $.795$ & \underline{$.759$} \\
\(A=0.25\) & \underline{$.925$} & $.822$ & \underline{$.193$} & $.298$ & $.791$ & \underline{$.767$} \\
\(A=0.35\) & \underline{$.926$} & $.820$ & \underline{$.178$} & $.288$ & $.778$ & \underline{$.764$} \\
\hline
\end{tabular}
\end{table}

\FloatBarrier
\appsubsection{Operating Regime of Evidence Transport}

\paragraph{Complete compact-plant matrix.}
The main text displays six of the eleven compact plants.  The complete
evaluation uses the settings in Table~\ref{tab:expM1_settings}: two unseen
gains, five seeds, 40 trials per seed and family, and all eight methods.
Table~\ref{tab:expM1_full_matrix} reports the rank of EG-MPT on all five
pre-specified metrics.  The point-estimation conclusion is broad but not
universal: compensation error ranks first on most plants, whereas interval
coverage does not.  The battery plant is the clearest exception, anticipating
the nonlinear-response failure mode isolated below.

\begin{table}[H]
\centering
\caption{Complete compact-plant result matrix behind
Table~\ref{tab:expM1_estimation}.  Each entry is the rank of EG-MPT among eight
methods over five seeds and two unseen severities.  A dagger marks a plant
shown in the main text.}
\label{tab:expM1_full_matrix}
\scriptsize
\setlength{\tabcolsep}{5pt}
\renewcommand{\arraystretch}{1.04}
\begin{tabular}{|l|c|c|c|c|c|}
\hline
Plant & Comp.\ error & Worst error & CRPS & Probes & Coverage \\
\hline
Attitude \(\dagger\) & \underline{$1$} & \underline{$1$} & $2$ & $2$ & $2$ \\
Building \(\dagger\) & \underline{$1$} & \underline{$1$} & $3$ & $2$ & $7$ \\
Pump network & \underline{$1$} & $2$ & $2$ & $2$ & $7$ \\
Battery pack & 5 & 4 & 3 & 2 & 7 \\
Reaction wheel & \underline{$1$} & $2$ & $2$ & $2$ & $2$ \\
Aircraft \(\dagger\) & \underline{$1$} & \underline{$1$} & $2$ & $2$ & $7$ \\
Distillation & \underline{$1$} & $2$ & $3$ & $2$ & $7$ \\
Servo \(\dagger\) & \underline{$1$} & \underline{$1$} & $2$ & $2$ & $5$ \\
Quad tank & $2$ & $2$ & $3$ & $2$ & $2$ \\
Quadrotor \(\dagger\) & \underline{$1$} & \underline{$1$} & $2$ & $2$ & $2$ \\
Turbine \(\dagger\) & \underline{$1$} & \underline{$1$} & \underline{$1$} & $2$ & $7$ \\
\hline
\end{tabular}
\end{table}

\begin{figure}[H]
\centering
\includegraphics[width=0.94\textwidth]{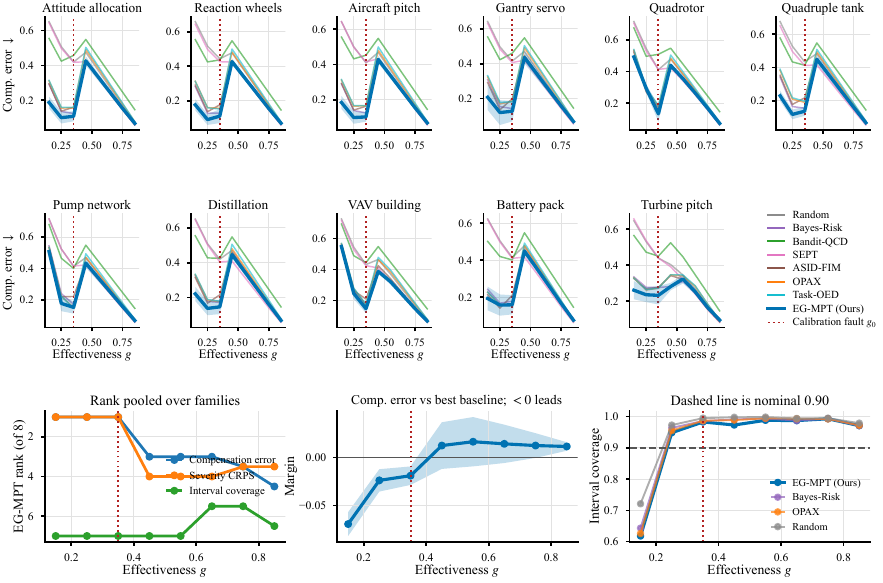}
\caption{Compensation error across eleven compact plants and a dense
effectiveness sweep.  The dotted line is the sole calibration fault
\(g_{\mathrm{cal}}=0.35\).  Summary panels report EG-MPT rank, its paired
margin to the best competitor, and interval coverage.}
\label{fig:expM2_severity_sweep}
\end{figure}

\paragraph{Dense severity sweep across compact plants.}
We next run all eleven families, all eight methods, five seeds, and 40 fault
trials per seed at
\(g\in\{0.15,0.25,0.35,0.45,0.55,0.65,0.75,0.85\}\), with a single
calibration anchor \(g_{\mathrm{cal}}=0.35\) and \(B_2=1\).
Figure~\ref{fig:expM2_severity_sweep} shows a sharp operating boundary:
transport leads for faults at or more severe than its calibration anchor, but
becomes mid-field for milder faults.  The matched response is larger than its
calibrated signature below the anchor and progressively noise-dominated above
it.  Calibration severity is therefore a design choice: it should be no more
severe than the mildest degradation for which continuous correction matters.

\begin{figure}[H]
\centering
\includegraphics[width=0.72\textwidth]{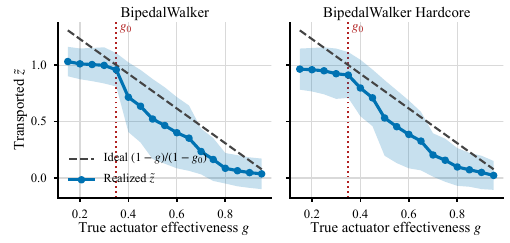}
\caption{Realized versus ideal transported coordinate on two Box2D backends.
The nominal and calibrated-fault anchors are exact; intermediate and
extrapolated severities expose bias, saturation, and heteroscedasticity.}
\label{fig:expN8_calibration}
\end{figure}

\paragraph{Calibration of the transported coordinate.}
On BipedalWalker-v3 and BipedalWalkerHardcore-v3, we evaluate the normalized
matched coordinate on the 17-point grid \(g=0.15,0.20,\ldots,0.95\), using
40 episodes per gain and mechanism and the fixed anchor
\(g_{\mathrm{cal}}=0.35\).  Figure~\ref{fig:expN8_calibration} verifies the
anchors but rejects the global linear assumption: the response saturates for
severe faults and its noise-to-signal ratio rises for mild faults.  This
explains why unseen-severity gains concentrate at the severe end and why a
constant transport-noise parameter can make a posterior sharp without making
it well calibrated.

\begin{figure}[H]
\centering
\includegraphics[width=0.72\textwidth]{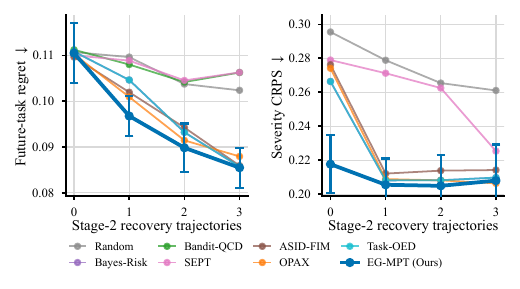}
\caption{Future-task regret and severity CRPS versus Stage~2 recovery budget,
macro-averaged over two Box2D backends and two unseen severities.  Seed
standard deviation is shown for EG-MPT.}
\label{fig:expN8_budget_curves}
\end{figure}

\paragraph{Equivalent recovery budget.}
Using the same two Box2D backends, we cross
\(g_{\mathrm{true}}\in\{0.20,0.50\}\) with
\(B_2\in\{0,1,2,3\}\), all eight methods, 50 trials per environment and seed,
and seeds \(0\)--\(4\).  Figure~\ref{fig:expN8_budget_curves} separates belief
quality from deployment value.  Transport improves severity CRPS even at
\(B_2=0\), but cannot reduce downstream regret until a recovery action is
available.  Once \(B_2\geq1\), the transported prior replaces a measurable
fraction of a physical recovery trajectory; as the budget grows, real
observations erase that advantage.

\begin{figure}[H]
\centering
\includegraphics[width=0.6\textwidth]{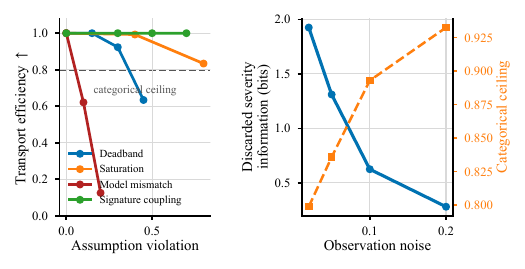}
\caption{Transport efficiency under controlled assumption violations (left)
and the disappearance of discarded severity information as observation noise
grows (right).}
\label{fig:mechanism_transport}
\end{figure}

\paragraph{Known-model mechanism study.}
To distinguish information present in the response from information recovered
by the estimator, we use an over-actuated attitude plant with a known
generative model.  The study uses five categorical bins, a 17-point gain grid,
120 Monte Carlo responses per gain, and independently varies deadband,
saturation, signature coupling, model mismatch, and observation noise.
Figure~\ref{fig:mechanism_transport} shows that transport reaches the Bayes
severity ceiling when its response model is correct.  Signature coupling harms
localization but not conditional severity recovery; deadband and saturation
break linearity; nominal-model mismatch is the most damaging violation.
Increasing observation noise is different: it removes the within-bin
information itself, leaving less for any transport rule to recover.

\FloatBarrier
\appsubsection{Component Attribution Across Plants and Power Systems}

\paragraph{Cross-family component ladder.}
We evaluate five rungs---full EG-MPT, no evidence gate, no transport, uniform
acquisition, and neither transport nor informed acquisition---while holding
the channel, prior, trajectory library, \(B_2=1\), correction, and deployment
rule fixed.  The synthetic ladder covers all eleven compact plants at gains
\(\{0.20,0.35\}\), 30 trials per family and seed, and seeds \(0\)--\(4\).
The Grid2Op companion uses the same rungs and gains, five seeds, and 10 trials
per seed.

\begin{figure}[H]
\centering
\includegraphics[width=0.94\textwidth]{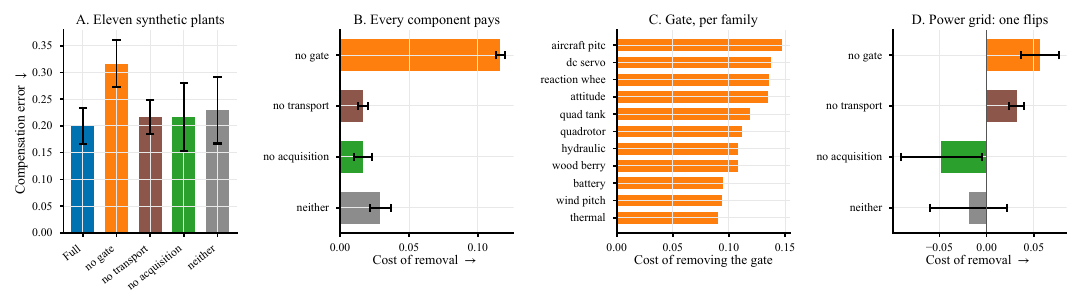}
\caption{Component ladder on eleven compact plants and Grid2Op.  Positive
removal cost means the removed component contributes to lower compensation
error.}
\label{fig:expG2_ladder}
\end{figure}

Figure~\ref{fig:expG2_ladder} identifies the evidence gate as the largest and
most consistent contributor.  Transport and informed acquisition are smaller:
their aggregate effects are positive, but a single benchmark often lacks the
resolution to separate them from zero.  On the power grid, uniform acquisition
appears better because generator signatures differ strongly and the greedy
selector concentrates on a subset of units.  The reversal is not resolved at
the available sample size, so it is a diagnostic of allocation imbalance, not
evidence that uniform probing is generally preferable.

\begin{figure}[H]
\centering
\includegraphics[width=0.80\textwidth]{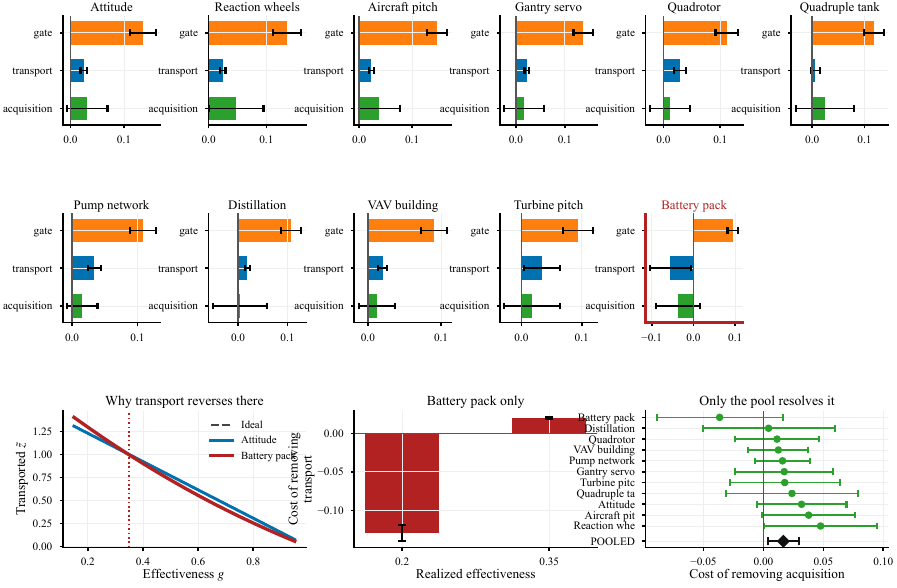}
\caption{Per-family cost of removing the gate, transport, or informed
acquisition.  The battery plant is highlighted as the one resolved reversal
for amplitude transport.}
\label{fig:expG3_per_family}
\end{figure}

\paragraph{Family-specific contribution and nonlinear failure.}
Figure~\ref{fig:expG3_per_family} resolves the pooled ladder by plant.  The gate
helps on every family, and transport helps on ten of eleven.  The battery pack
is the exception: saturation and state-of-charge-dependent response bend the
matched coordinate to opposite sides of the assumed linear relationship.
Transport then sharpens the posterior around a biased severity.  Acquisition,
by contrast, remains too small to resolve on any single family and becomes
detectable only after pooling the family--severity--seed cells.

\begin{figure}[H]
\centering
\includegraphics[width=0.84\textwidth]{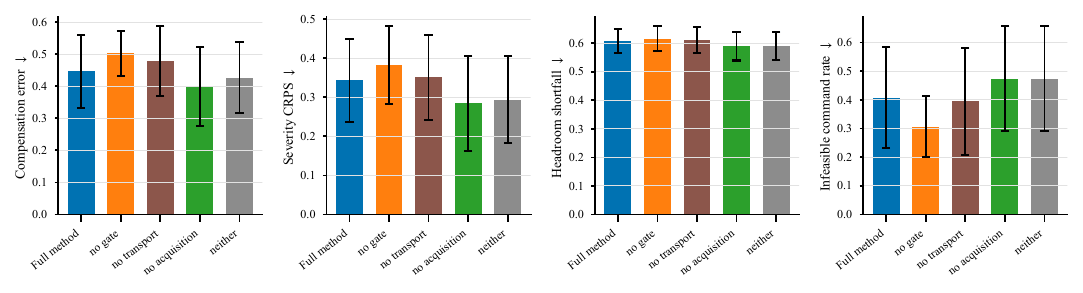}
\caption{Component ladder on Grid2Op, including posterior metrics and
grid-native headroom shortfall and ramp-infeasibility rate.}
\label{fig:expG1_ladder}
\end{figure}

\paragraph{Grid2Op actuator limits.}
The Grid2Op L2RPN NeurIPS-2020 study uses six redispatchable generators: the
largest-ramp unit balances five dormant candidates whose ramp capabilities
span a factor of roughly seven.  Diagnostics are eight-step zero-sum
redispatch pulses, calibrated at \(g=0.35\) with 20 episodes and five bins;
the two true gains are \(0.20\) and \(0.35\), with \(B_2=1\).
Figure~\ref{fig:expG1_ladder} confirms that the gate and transport improve
severity estimation, but it also exposes a control constraint absent from the
posterior model.  A correction for severe loss can request more redispatch than
a generator can physically ramp, so clipping dominates headroom shortfall and
compresses differences between rungs.  A deployable readiness certificate
must therefore propagate actuator limits, not only uncertainty in \(g\).

\FloatBarrier
\appsubsection{Domain-Native Control and Safety}

\paragraph{Directional control without recovery interaction.}
CartPole-v1 uses right and left directional actuation as dormant mechanisms.
The right diagnostic has cost 0.01, the noisier left diagnostic has cost 0.05,
future-task mass is \(0.4/0.6\), and relative importance is \(1.0/2.5\).
Pendulum-v1 uses the same directional task weighting without the nuisance
term.  Both use calibration gain \(0.35\), \(B_2=0\), 30 fault trials per seed,
and seeds \(0\)--\(4\); Table~\ref{tab:cartpole_directional_allocation}
reports gains \(0.25\) and \(0.35\).

The severe CartPole setting benefits strongly from transporting within-bin
amplitude, while the milder setting is much closer to the strongest
decision-theoretic baselines.  Pendulum shows the same direction with a smaller
margin.  Because no recovery rollout is available, these differences arise
from Stage~1 evidence and the resulting fallback/deployment choice rather than
from additional system identification.

\begin{table}[H]
\centering
\caption{Directional control with no Stage~2 recovery interaction.  CartPole
reports stability regret and Pendulum reports return regret; lower is better.
Entries are mean \(\pm\) sample standard deviation over five seeds.}
\label{tab:cartpole_directional_allocation}
\scriptsize
\setlength{\tabcolsep}{3.8pt}
\renewcommand{\arraystretch}{1.04}
\begin{tabular}{|l|c|c|c|c|}
\hline
& \multicolumn{2}{c|}{CartPole-v1}
& \multicolumn{2}{c|}{Pendulum-v1} \\
\hline
Method & \(g=.25\) & \(g=.35\) & \(g=.25\) & \(g=.35\) \\
\hline
Random & \(430.3\pm28.4\) & \(123.9\pm32.2\)
& \(55.35\pm23.28\) & \(61.27\pm18.28\) \\
Bayes-Risk & \(425.7\pm28.0\) & \(37.8\pm22.4\)
& \(61.84\pm16.33\) & \(63.44\pm15.36\) \\
\hline
Bandit-QCD & \(440.0\pm29.2\) & \(432.9\pm29.3\)
& \(89.06\pm27.23\) & \(100.94\pm28.18\) \\
SEPT & \(428.8\pm28.6\) & \(40.9\pm28.7\)
& \(62.41\pm16.74\) & \(62.96\pm13.87\) \\
ASID-FIM & \(429.3\pm27.9\) & \(59.2\pm28.4\)
& \(68.12\pm15.08\) & \(70.25\pm15.02\) \\
OPAX & \(429.3\pm27.9\) & \(59.2\pm28.4\)
& \(61.84\pm16.33\) & \(63.44\pm15.36\) \\
Task-OED & \(425.7\pm28.0\) & \(37.8\pm22.4\)
& \(61.84\pm16.33\) & \(63.44\pm15.36\) \\
\hline
\textbf{EG-MPT (Ours)} & \underline{$136.0\pm42.9$} & \underline{$32.2\pm16.5$}
& \underline{$40.89\pm10.34$} & \underline{$44.32\pm23.51$} \\
\hline
\end{tabular}
\end{table}

\paragraph{Driving safety on highway-env.}
In highway-v0, throttle is the current mechanism and braking, left steering,
and right steering are dormant.  Diagnostics use 16-step zero-integral pulses;
nominal and calibrated references are cached separately for each traffic
scene.  We test gains \(\{0.20,0.35\}\), calibration gain \(0.35\),
\(B_2=1\), 15 trials per method and seed, and seeds \(0\)--\(4\).
Evaluation uses native safety quantities: collision rate, minimum time to
collision, minimum headway, and lower-tail time to collision.

\begin{figure}[H]
\centering
\includegraphics[width=1.0\textwidth]{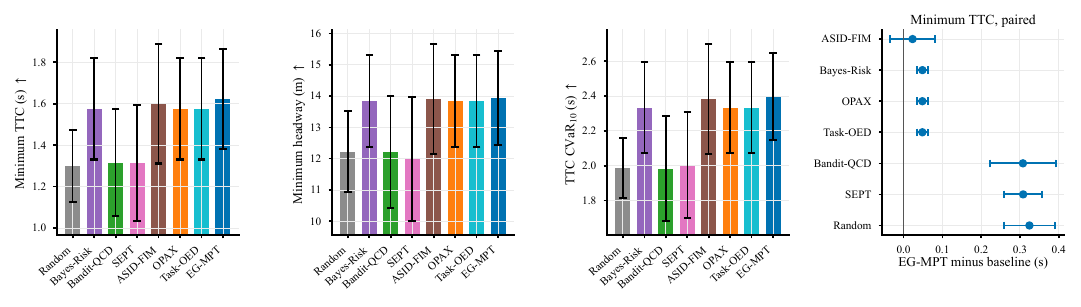}
\caption{Native driving-safety metrics on highway-env at two unseen
severities.  The paired panel reports EG-MPT's minimum-time-to-collision
advantage over each baseline.}
\label{fig:expH1_safety}
\end{figure}

Figure~\ref{fig:expH1_safety} shows that EG-MPT improves safety margins against
most baselines while tying the best collision rate.  ASID-FIM is statistically
indistinguishable at this sample size.  The insight is therefore about margin,
not crash-count dominance: better severity-conditioned correction creates more
time and space before failure, but the binary collision endpoint is too sparse
to separate the strongest methods.

\FloatBarrier
\appsubsection{Non-Mechanical Generalization Control}

\paragraph{Supplier-yield readiness in GymInvMgmt.}
We treat replenishment links as mechanisms whose delivered order is multiplied
by an unknown supplier-yield gain.  The completed matrix contains the
GymInvMgmt Serial backlog, Serial lost-sales, and MultiEchelon environments;
all eight methods; gains \(\{0.40,0.60,0.80\}\); and a demand-shock setting
with \(g=0.60\) and doubled demand stress.  Each job uses a 60-period
deployment, \(B_2=1\), 30 trials per seed, and seeds \(0\)--\(4\).
Table~\ref{tab:inventory_control} reports metrics native to inventory control
rather than trajectory tracking.

The result is a useful boundary rather than an additional win.  EG-MPT is
close to the strongest baseline on on-time-in-full service, but does not
consistently improve safe contract coverage and produces more false-ready
contracts.  A multiplicative supplier-yield fault enters the order action
simply, yet its service consequence is filtered through lead times, backlog,
order clipping, and the inventory policy.  A matched one-step response is
therefore not a sufficient statistic for downstream readiness.  Applying the
method outside actuator-effectiveness problems requires a domain-specific
response model and certificate, not merely relabeling a supply link as an
actuator.

\begin{table}[H]
\centering
\caption{GymInvMgmt generalization control, macro-averaged over three
environments and five seeds.  ``Best base'' is the strongest baseline for each
metric and setting.}
\label{tab:inventory_control}
\scriptsize
\setlength{\tabcolsep}{4.0pt}
\renewcommand{\arraystretch}{1.05}
\begin{tabular}{|l|c|c|c|c|c|c|}
\hline
& \multicolumn{2}{c|}{OTIF \(\uparrow\)}
& \multicolumn{2}{c|}{Safe coverage \(\uparrow\)}
& \multicolumn{2}{c|}{False ready \(\downarrow\)} \\
\hline
Setting & EG-MPT & Best base & EG-MPT & Best base & EG-MPT & Best base \\
\hline
Demand shock & $.763$ & \underline{$.773$} & $.107$ & \underline{$.155$} & $.078$ & \underline{$.000$} \\
\(g=0.40\) & $.865$ & \underline{$.866$} & \underline{$.352$} & $.351$ & $.041$ & \underline{$.011$} \\
\(g=0.60\) & $.842$ & \underline{$.854$} & $.140$ & \underline{$.218$} & $.044$ & \underline{$.000$} \\
\(g=0.80\) & $.842$ & \underline{$.849$} & $.129$ & \underline{$.209$} & $.056$ & \underline{$.000$} \\
\hline
\end{tabular}
\end{table}

\newpage
\appsection{Extended Discussion}
\label{app:extended_discussion}

\appsubsection{Potential Impact}

This work treats readiness as a decision made under hidden dynamics changes,
rather than as change detection alone.  The practical benefit is a more
disciplined use of limited interaction: targeted diagnostics localize the
affected mechanism, transported evidence supports severity-aware recovery, and
task-wise certificates expose when deployment is not justified.  Such a
pipeline may help reduce unsafe trial-and-error after actuator degradation in
robotics, industrial control, energy systems, and other settings in which
online experimentation is costly.  Just as importantly, explicit abstention
can make uncertainty visible instead of silently converting a weak diagnosis
into an aggressive control action.

These benefits do not by themselves establish safety.  A readiness certificate
is only as meaningful as its task library, loss model, calibration data, and
fallback policy.  In high-consequence applications, EG-MPT should therefore be
used as one component of a broader assurance process that includes physical
constraints, independent monitoring, operator oversight, and domain-specific
validation.

\appsubsection{Limitations and Scope}

The method is designed for localized, intervention-responsive dynamics changes
with diagnostic signatures represented in the calibration channel.  Its
matched-response transport is most reliable when response amplitude varies
approximately monotonically with mechanism effectiveness.  Saturation,
deadbands, strong mechanism coupling, model mismatch, or state-dependent
effects can violate this structure; the additional experiments show that
transport may then become biased rather than merely less informative.
Likewise, weak faults near the noise floor cannot be recovered by reusing
evidence that the diagnostic never contained.

Our certificates are empirical and distribution-dependent, not formal
closed-loop safety guarantees.  They cover a finite task and correction
library, and their conclusions need not transfer to unseen objectives,
operating regions, simultaneous faults, or adversarial changes.  The present
evaluation also emphasizes simulated benchmarks and compact controlled
systems.  Larger physical platforms may introduce delayed observations,
nonstationary noise, hard actuator limits, and intervention costs that are not
fully captured here.

\appsubsection{Future Directions}

A natural next step is to replace a fixed calibration channel with adaptive
response models that detect when transport assumptions fail and revert to
conservative recovery.  Joint inference over multiple interacting faults,
continuous task representations, and diagnostics optimized under explicit
safety constraints would broaden the formulation.  Another important
direction is to propagate actuator limits and model uncertainty through the
certificate itself, so that posterior confidence cannot mask infeasible
corrections.  Finally, hardware studies with prespecified fallback rules and
out-of-distribution audits are needed to determine when the observed gains in
sample-efficient readiness translate into dependable real-world operation.

\appsubsection{Code and Reproducibility}

Upon acceptance, we will publicly release the complete research artifacts
needed to reproduce all reported results.  The release will include source
code, environment and experiment configurations, data-generation and
evaluation scripts, random seeds, model checkpoints, raw logs, and plotting
utilities, together with instructions for regenerating every table and figure
in the paper.  These materials are not being distributed during the current
review stage in order to preserve anonymity and avoid unintended circulation
before publication.  The accepted-paper release will be versioned and archived
at a persistent public location so that the reported experiments can be
independently audited and reproduced.

\end{document}